\pdfoutput=1

\documentclass[11pt]{article}
\usepackage{CJKutf8}
\usepackage{acl}

\usepackage{times}
\usepackage{latexsym}

\usepackage[T1]{fontenc}

\usepackage[utf8]{inputenc}

\usepackage{microtype}

\usepackage{inconsolata}

\usepackage{times}
\usepackage{latexsym}
\usepackage{balance}
\usepackage[T1]{fontenc}

\usepackage[utf8]{inputenc}

\usepackage{microtype}

\usepackage{inconsolata}
\usepackage{CJKutf8}
\usepackage{booktabs}
\usepackage{graphicx}
\usepackage{array}
\usepackage{bm}
\usepackage{amsmath}
\usepackage{enumitem}
\usepackage{threeparttable}
\usepackage{multirow}
\usepackage{makecell}

\usepackage{latexsym} 

\usepackage[T1]{fontenc}

\usepackage[utf8]{inputenc}
\usepackage{float}

\usepackage{microtype}

\usepackage{xcolor}

\def\BibTeX{{\rm B\kern-.05em{\sc i\kern-.025em b}\kern-.08em
    T\kern-.1667em\lower.7ex\hbox{E}\kern-.125emX}}
\usepackage{cite}
\usepackage{amsmath,amssymb,amsfonts}
\usepackage{algorithmic}
\usepackage[ruled,linesnumbered,noline,noend]{algorithm2e}

\SetCommentSty{mycommfont}

\usepackage[medium,compact]{titlesec}

\SetNlSty{}{}{}

\let\oldnl\nl
\newcommand\nonl{%
  \renewcommand{\nl}{\let\nl\oldnl}}
  
\usepackage{hyperref}
\hypersetup{
    colorlinks=true,
    linkcolor=blue,}
\AtBeginDocument{%
  \providecommand\BibTeX{{%
    \normalfont B\kern-0.5em{\scshape i\kern-0.25em b}\kern-0.8em\TeX}}}

\usepackage{textcomp}
\usepackage{xcolor}
\def\BibTeX{{\rm B\kern-.05em{\sc i\kern-.025em b}\kern-.08em
    T\kern-.1667em\lower.7ex\hbox{E}\kern-.125emX}}
\usepackage{amsmath,amssymb,amsfonts}
\usepackage{algorithmic}
\usepackage[ruled,linesnumbered,noline,noend]{algorithm2e}
\usepackage{threeparttable}
\usepackage{booktabs}
\usepackage{graphicx}
\usepackage{textcomp}
\usepackage{array}
\usepackage{bm}
\usepackage{amsmath}
\usepackage{color, colortbl}
\SetCommentSty{mycommfont}

\usepackage[misc]{ifsym}
\usepackage{amsthm}
\newtheorem{Problem definition}{Problem definition}

\def\BibTeX{{\rm B\kern-.05em{\sc i\kern-.025em b}\kern-.08em
    T\kern-.1667em\lower.7ex\hbox{E}\kern-.125emX}}
\usepackage{hyperref}
\hypersetup{
    colorlinks=true,
    linkcolor=blue,}

\usepackage{color, colortbl}
\usepackage{array}
\usepackage{bm}
\usepackage{amsmath}
\usepackage{threeparttable}
\usepackage{booktabs}
\usepackage{graphicx}
\usepackage{hyperref}
\usepackage{multirow}
\usepackage[ruled,linesnumbered,noline,noend]{algorithm2e}
\usepackage{makecell}
\hypersetup{
    colorlinks=true,
    linkcolor=blue,}
\usepackage{CJKutf8}
\usepackage{booktabs}
\usepackage{graphicx}
\usepackage{textcomp}
\usepackage{array}
\usepackage{bm}
\usepackage{amsmath}

\usepackage{threeparttable}
\usepackage{multirow}
\usepackage{makecell}
\usepackage{times}
\usepackage{latexsym}
\usepackage[T1]{fontenc}

\usepackage[utf8]{inputenc}
\usepackage{float}

\usepackage{microtype}

\usepackage{xcolor}

\def\BibTeX{{\rm B\kern-.05em{\sc i\kern-.025em b}\kern-.08em
    T\kern-.1667em\lower.7ex\hbox{E}\kern-.125emX}}
\usepackage{cite}
\usepackage{amsmath,amssymb,amsfonts}
\usepackage{algorithmic}
\usepackage[ruled,linesnumbered,noline,noend]{algorithm2e}

\SetCommentSty{mycommfont}

\usepackage{fontawesome}

\SetNlSty{}{}{}

\let\oldnl\nl

\usepackage{hyperref}
\hypersetup{
    colorlinks=true,
    linkcolor=blue,}
\AtBeginDocument{%
  \providecommand\BibTeX{{%
    \normalfont B\kern-0.5em{\scshape i\kern-0.25em b}\kern-0.8em\TeX}}}
\usepackage[medium,compact]{titlesec}

\usepackage{times}
\usepackage{latexsym}

\usepackage[T1]{fontenc}

\usepackage[utf8]{inputenc}

\usepackage{microtype}
\usepackage{mathrsfs}

\title{Dr.Academy: A Benchmark for Evaluating Questioning Capability in Education for Large Language Models}

\author{Yuyan Chen$^{1}$, Chenwei Wu$^{2}$, Songzhou Yan$^{1}$, Panjun Liu$^{3}$, Haoyu Zhou,\textbf{Yanghua Xiao}$^{1}$ $^{\textrm{\Letter}}$\\
        $^1$Shanghai Key Laboratory of Data Science, School of Computer Science, Fudan University, \\
        $^2$Electrical Engineering and Computer Science Department, University of Michigan,\\
        $^3$School of Computer Science, Beijing Institute of Technology
        \\
        \texttt{\{chenyuyan21@m., szyan21@m., shawyh@\}fudan.edu.cn},\\
        \texttt{chenweiw@umich.edu},
        \texttt{panjunliu@outlook.com},
        \texttt{haoyucharleszhou@gmail.com}
        }

\begin{document}
\maketitle
\begin{abstract}
Teachers are important to imparting knowledge and guiding learners, and the role of large language models (LLMs) as potential educators is emerging as an important area of study. Recognizing LLMs' capability to generate educational content can lead to advances in automated and personalized learning. While LLMs have been tested for their comprehension and problem-solving skills, their capability in teaching remains largely unexplored.
In teaching, questioning is a key skill that guides students to analyze, evaluate, and synthesize core concepts and principles.
Therefore, our research introduces a benchmark to evaluate the questioning capability in education as a teacher of LLMs through evaluating their generated educational questions, utilizing Anderson and Krathwohl's taxonomy across general, monodisciplinary, and interdisciplinary domains. We shift the focus from LLMs as learners to LLMs as educators, assessing their teaching capability through guiding them to generate questions. We apply four metrics, including relevance, coverage, representativeness, and consistency, to evaluate the educational quality of LLMs' outputs. Our results indicate that GPT-4 demonstrates significant potential in teaching general, humanities, and science courses; Claude2 appears more apt as an interdisciplinary teacher. Furthermore, the automatic scores align with human perspectives.
\end{abstract}

\section{Introduction}

Large language models (LLMs) have demonstrated great performance in various natural language processing (NLP) tasks, including question answering~\citep{saad2023pdftriage,chen2024temporalmed,chen2024talk,chen2024xmecap}, information retrieval~\citep{chen2022grow,liu2023reta}, reasoning~\citep{kojima2022large,chen2023can}, and generation~\citep{chung2023increasing,chen2023xmqas,chen2024hotvcom}, etc. Beyond these general NLP applications, LLMs are also widely used in other domains, such as education. In the educational field, LLMs can now be used as substitutes for teachers. They can help automated teaching or assisted learning applications, thereby alleviating the pressure on human teachers. Additionally, LLMs can recommend appropriate elective courses based on a student's knowledge state, learning style, and interests, automatically generating practice problems of corresponding difficulty levels, and identifying areas where a student is struggling to provide targeted improvement.

\begin{figure}[t]
  \centering
  \includegraphics[width=0.9\linewidth]{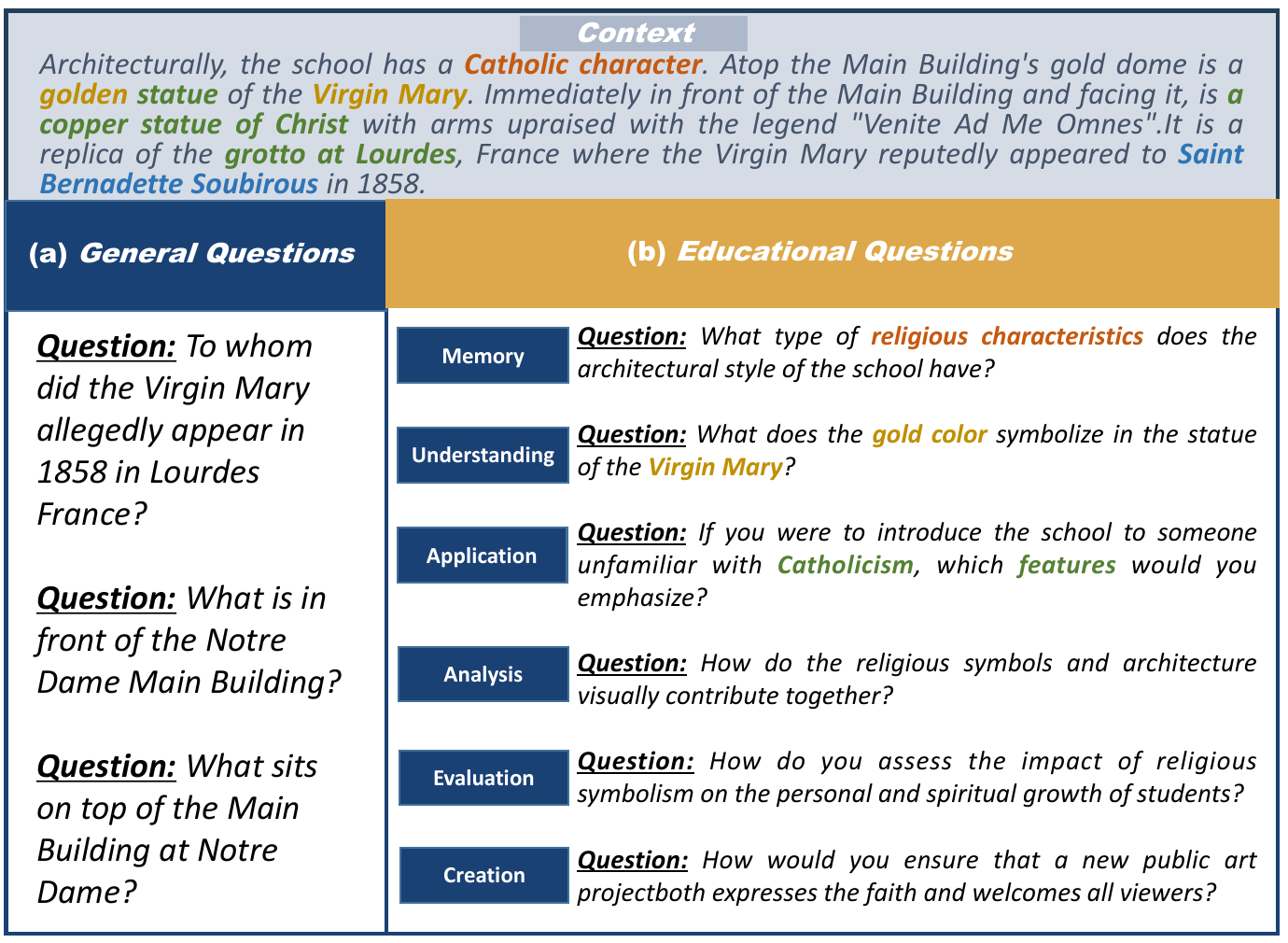}
  \caption{Comparison between general and educational questions.}
  \label{fig:xuexi-intro}
\end{figure}

However, the capability of questioning is a crucial aspect in the educational field. As LLMs take on the role of teachers, can they pose high-quality questions like human educators? Therefore, evaluating what constitutes a high-quality question in education becomes necessary. According to Anderson and Krathwohl's educational taxonomy~\citep{anderson2001taxonomy,elkins2023useful}, we consider that high-quality questioning in the educational field must meet the following characteristics: i) achieve a higher level across the six domains including memory, understanding, application, analysis, evaluation, and creation; ii) be relevant to the given context; iii) comprehensively cover the content of the context, and iv) also reflect the important knowledge of this context. We consider that questions meeting these characteristics can effectively assess students' knowledge levels, and LLMs capable of posing such questions can assume the role of competent human educators. The first characteristic is the most basic requirement for LLMs to act as human teachers, while the following three characteristics measure the excellence of LLMs in their role as a teacher.

Evaluating and enhancing the capability of LLMs to generate questions of high quality standards in the educational domain requires a benchmark.
However, previous studies have mainly viewed LLMs from a student's perspective, focusing on tasks like reading comprehension~\citep{bai2023longbench,tran2023single,chen2023modeling,cheng2023adapting,izacard2020leveraging,kawabata2023evaluating,zhou2023thread,zhou2023complementary} and exam evaluations~\citep{zhang2023evaluating,huang2023c,zhong2023agieval,li2023cmmlu,wang2023scibench,zeng2023measuring,wei2023cmath}. 
However, these tasks focus on adopting contexts to passively answer questions or make reasoning, and these tests treat LLMs as students, assessing their abilities by how they answer questions, while the LLM's questioning capability through generating educational questions is under-studied. Current education-related research is far from adequate to determine LLMs' question raising capability as a teacher, and there isn't a benchmark that studies the overall teaching abilities of LLMs, seeing them as teachers.
Although some role-playing tasks~\citep{shao2023character} mimic professional dialogues but don't truly assess the LLMs' teaching capabilities. Therefore, if we want LLMs to assist in teaching effectively, we need to evaluate and enhance their teaching abilities, as possessing knowledge and guiding others to learn are distinct skills.

Therefore, in this paper, we have developed a benchmark for assessing whether LLMs generate high-quality questions in the field of education, guided by professional educational theories. Unlike general questioning, as shown in Fig.~\ref{fig:xuexi-intro} (a), our benchmark requires that the generated questions not only be fluent and readable but also meet the fundamental characteristics proposed earlier (i.e. the first characteristic), as shown in Fig.~\ref{fig:xuexi-intro}(b). Specifically, we draw on Anderson and Krathwohl's educational taxonomy~\citep{anderson2001taxonomy} to prompt LLMs to generate questions at six levels for each context. We select tasks from three domains, including general, single-discipline, and interdisciplinary domains, to more comprehensively assess the strengths of LLMs as teachers in various fields. Based on the four characteristics proposed earlier, we have also designed four evaluation metrics: consistency, relevance, coverage, and representativeness, to assess the value of questions posed by LLMs in the educational domain, thereby comprehensively evaluating the questioning capability of LLMs as teachers in education through evaluating their generated educational questions. Our experiments reveal that LLMs like GPT-4, Claude2, and GPT-3.5 demonstrate good questioning capability across domains as teachers in education through evaluating their generated educational questions.
In summary, our contributions are threefolds:
\begin{itemize}
    \item We introduce the problem of evaluating questioning capability in education as a teacher for LLMs through evaluating their generated educational questions, building a framework based on educational theory that includes six cognitive levels and tasks from three different domains.
    \item We establish four evaluation metrics to assess the questioning capability in education as a teacher of LLMs through evaluating their generated educational questions.
    \item We conduct experimental evaluations of 11 LLMs, providing quantitative standards and subject orientations for each LLM's questioning capability as a teacher.
\end{itemize}

\section{Datasets and Task Setups}

\begin{figure*}[!h]
  \centering
  \includegraphics[width=0.95\linewidth]{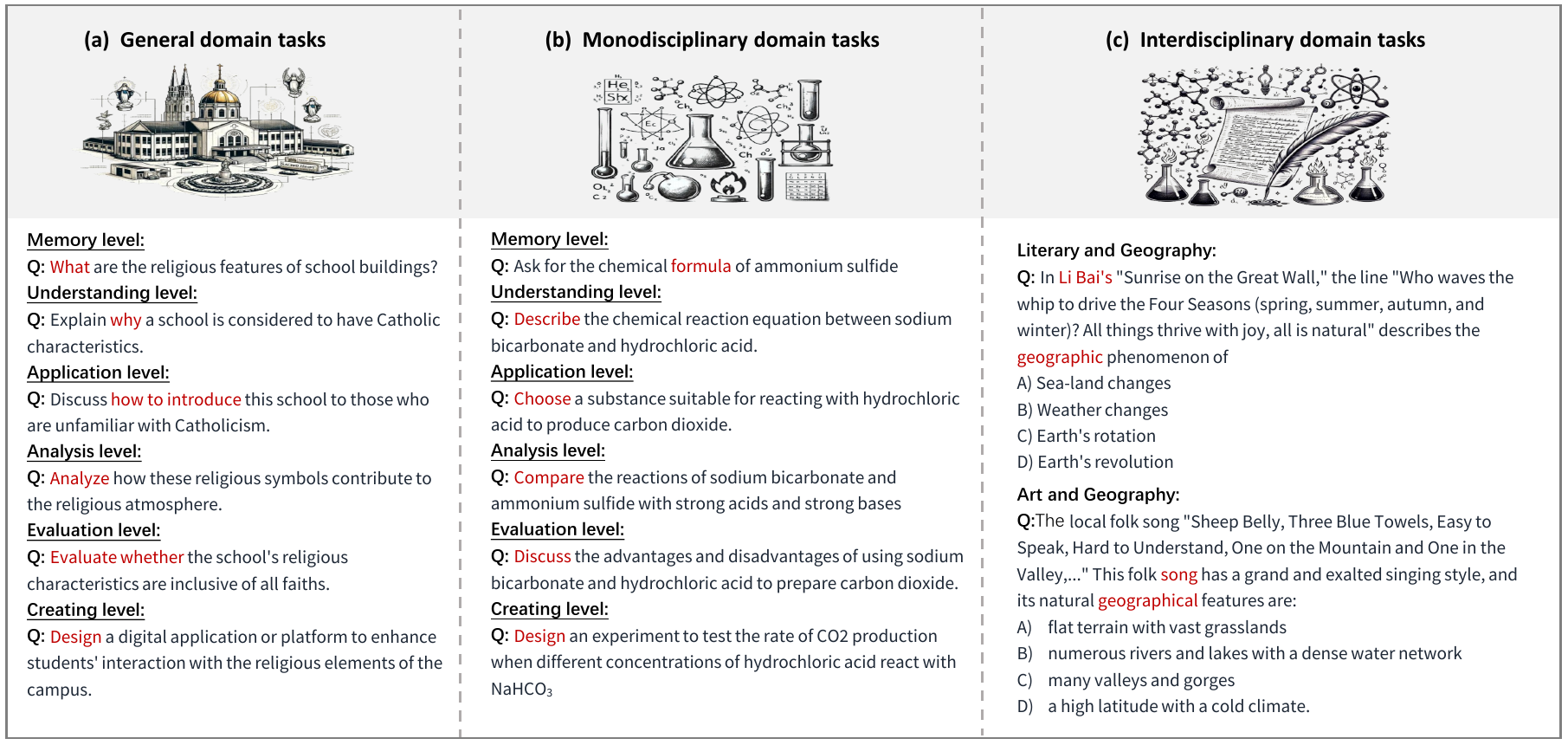}
  \caption{The overview of the proposed benchmark Dr.Academy, including three domains and different levels.}
  \label{fig:xuexi-task}
\end{figure*}

We propose a benchmark named Dr.Academy, which has tasks from three domains.
The first two request LLMs to generate questions in the general and monodisciplinary domain, respectively, based on the six levels of Anderson and Krathwohl's educational taxonomy~\citep{anderson2001taxonomy}, including memory, understanding, application, analysis, evaluation and creation. The third one requests LLMs to generate questions that intersect multiple subjects.
The overview of Dr.Academy is shown in Fig.~\ref{fig:xuexi-task}.

\subsection{Context Construction}
Initially, we collect 10,000 contexts from the general domain and produce an additional 10,000 contexts specifically for the monodisciplinary domain. In the general domain, the contexts are sourced from the SQuAD dataset~\citep{rajpurkar2016squad}, an extractive reading comprehension dataset derived from Wikipedia articles, and are utilized as the foundation for the LLMs to generate questions.
In the monodisciplinary domain, we generate corresponding contexts for each of the multiple-choice questions in the MMLU dataset~\citep{hendrycks2020measuring}, which covers a broad spectrum of subjects, with GPT4~\footnote{https://chat.openai.com/}. These contexts include essential information related to the question and all candidate choices. The prompt for generating contexts is shown in Table~\ref{tab:xuexi-context-prompt}.
We also conduct manual evaluations on the generated contexts for the MMLU questions. In this process, we engage three graduate students from different disciplines to perform the evaluations. For each discipline, we randomly select 1\% of the questions to undergo manual assessment. If these entries do not achieve a manual evaluation score of 4 (on a scale of 1-5), we will regenerate the contexts.

\begin{table*}[]
\centering
\resizebox{\textwidth}{!}{%
\begin{tabular}{p{26cm}}
\toprule
Prompt                                                                                                                                        \\ \midrule
Craft a comprehensive and integrative textbook text that seamlessly weaves together the various knowledge points encompassed by the multiple-choice options. Begin by presenting a thorough exploration of each topic in individual paragraphs, ensuring the narrative flows logically from one subject to the next. The text should serve as a foundation that could logically precede the multiple-choice question. The goal is to offer a well-rounded and complete context that encapsulates the knowledge points mentioned in the options, setting the stage for the question to arise naturally from the presented material. \\ \bottomrule
\end{tabular}%
}
\caption{The instruction prompt in generating contexts based on a given multiple-choice questions.}
\label{tab:xuexi-context-prompt}
\end{table*}

\subsection{Task Setup}
We have designed three tasks and each task requires LLMs to generate questions catering for the corresponding domain. Finally, these generated questions will be used to evaluate the questioning capability in education as a teacher of LLMs. The prompt for generating questions is shown in Table~\ref{tab:xuexi-prompt} (row ``Generation'').

\textbf{General domain tasks} request an LLM to generate questions with the collected contexts from SQuAD based on the six levels of Anderson and Krathwohl's educational taxonomy~\citep{anderson2001taxonomy}, including memory, understanding, application, analysis, evaluation and creation. For instance, in Fig.~\ref{fig:xuexi-task} (a), at the memory level, a question might ask specific details like ``\emph{What are the religious features of a school building?}''; at the understanding level, it could be about reasons such as ``\emph{Why is a school considered to have Catholic characteristics?}''; at the creating level, questions could be more open-ended, involving imagination and design, etc. This task is designed to evaluate ``which LLM is more suitable to be a general course teacher''.

\textbf{Monodisciplinary domain tasks} request an LLM to generate questions with the generated contexts from MMLU, focusing on either humanities (like history, geography) or sciences (like physics, chemistry), based on the same six educational levels. 
In science, for instance, a memory-level question might ask about element symbols and formulas, such as ``\emph{What is the chemical formula for ammonium sulfate?}''; an application-level question related to real-world phenomena, like ``\emph{choose a substance that reacts with hydrochloric acid to produce carbon dioxide}''. This task is designed to evaluate ``which of the two LLMs is more suitable to act as a humanities teacher and a science teacher.''

\textbf{Interdisciplinary domain tasks} request an LLM to generate questions that cross multiple subject areas, reflecting each subject's characteristics. For example, in Fig.~\ref{fig:xuexi-task}(c), when merging literature and geography, a question might seek an explanation of the geographical phenomenon described in a poem's line. In combining art and geography, a question might ask about the geographical features represented in a song. 
A less successful example of an interdisciplinary question is one where the involved disciplines are unrelated, such as asking about Einstein's theory of relativity and then about the Cauchy inequality in mathematics. This question touches on physics and mathematics but lacks a meaningful connection between the two, making it not truly interdisciplinary.
This task is designed to evaluate ``which LLM is more suitable to act as a interdisciplinary teacher.'', qualifying if LLMs solving the problem requires understanding knowledge from both subjects.

\subsection{Evaluation metrics}

We adopt consistency, relevance, coverage, and representativeness to evaluate LLMs' generated questions in the general and monodisciplinary domains, respectively, while using relevance and representativeness to evaluate questions in the interdisciplinary domain. 
The difference of metrics selection is because questions in interdisciplinary domain lack a comprehensive contextual framework. For instance, in reality, there is no distinct academic discipline like ``historical geography.'' This absence of a well-defined, unified context means that metrics such as coverage and consistency do not apply.

\begin{figure}[!h]
  \centering
  \includegraphics[width=0.95\linewidth]{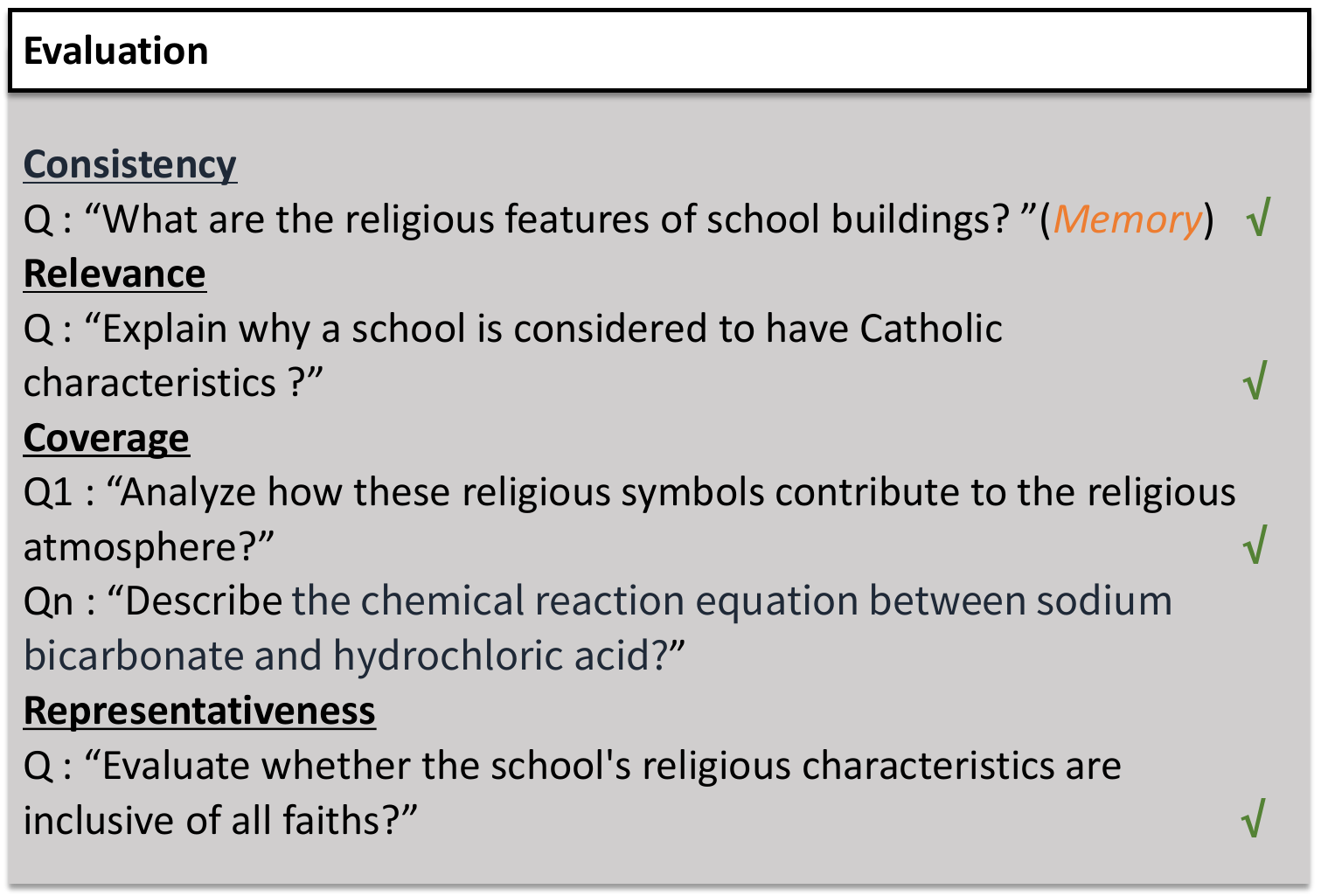}
  \caption{The evaluation metrics for assessing the quality of generated questions.}
  \label{fig:xuexi-evaluation}
\end{figure}

To validate the effectiveness of these metrics, we consult ten experts in education to rate the effectiveness of these metrics within the field of education on a scale of 1 to 5. They consistently award these metrics scores of 4 and above, which leads us to believe that these metrics are meaningful for evaluating questions in education. We also align these metrics with manual evaluations in Figure 6, indicating that our metrics are indeed significant within the field of education. 
Additionally, experiments are conducted to compare these metrics with human scoring in order to corroborate the validity and reasonableness of them (see Fig.~\ref{fig:xuexi-exp3}).
The prompt for evaluating questions is shown in Table~\ref{tab:xuexi-prompt} (see row ``Evaluation'').
Specifically, consistency is to assess whether the question accurately corresponds to a pre-defined level of the educational taxonomy,
relevance is to assess whether the question is related to the provided text content or theme,
coverage is assessed by determining if all generated questions based on a context encompasses a major portion (over 50\%) of this given context, 
representativeness evaluates whether a question captures the main content or core ideas of the text. 
Metrics are rated on a binary scale, with 1 for criteria met and 0 for not met, as shown in Fig.~\ref{fig:xuexi-evaluation} and Table~\ref{tab:criteria}. We adopt GPT-4 to score each question three times. A question that scores 1 in two out of three instances meets the metric's requirement~\citep{chen2023hallucination,chen2023mapo}.

\begin{table*}[]
\centering
\resizebox{\textwidth}{!}{%
\begin{tabular}{p{3cm}|p{22cm}|p{1cm}}
\toprule
Metric                              & Criteria                                                                                                                                                                                                             & Score \\ \midrule
\multirow{2}{*}{Consistency}        & The question does not align with any of the six educational levels (memory, understanding, application, analysis, evaluation, creation) as outlined in Anderson and Krathwohl's revised version of Bloom's Taxonomy. & 0     \\
                                    & The question aligns with one of the six educational levels (memory, understanding, application, analysis, evaluation, creation) as mentioned in Anderson and Krathwohl's revised version of Bloom's Taxonomy.        & 1     \\\midrule
\multirow{2}{*}{Relevance}          & The question is no more than 50\% relevant to the provided textual context.                                                                                                                                          & 0     \\
                                    & The question is more than 50\% relevant to the provided textual context.                                                                                                                                             & 1     \\\midrule
\multirow{2}{*}{Coverage}           & The content addressed by all questions does not cover more than 50\% of the textual context.                                                                                                                         & 0     \\
                                    & The content addressed by all questions covers more than 50\% of the textual context.                                                                                                                                 & 1     \\\midrule
\multirow{2}{*}{Representativeness} & The question does not represent more than 50\% of the important content within the context.                                                                                                                          & 0     \\
                                    & The question represents more than 50\% of the important content within the context.                                                                                                                                  & 1     \\\midrule 
\end{tabular}%
}
\caption{Rating criteria of each metric for evaluating generated questions.}
\label{tab:criteria}
\end{table*}

\section{Experiments}
In this section, we conduct extensive experiments to evaluate different LLMs' questioning capability through evaluating their generated educational questions in the proposed Dr.Academy.

\subsection{Experimental Setups}
Our experiments are conducted on 8 Nvidia A100 GPUs, each with 80GB of memory, and we use PyTorch~\footnote{https://pytorch.org/} in Python~\footnote{https://www.python.org/}. We set the maximum sequence length for both input and output sequences to maximum 1000 tokens.

\subsection{Datasets, Baselines and Metrics}
The baseline LLMs for this evaluation are BLOOM-7B~\citep{workshop2023bloom}
BLOOM-176B~\citep{workshop2023bloom},
Claude2~\citep{bai2022constitutional},
Falcon-7B~\citep{almazrouei2023falcon},
Falcon-180B~\citep{almazrouei2023falcon},
GPT3.5~\citep{brown2020language}, 
GPT4~\citep{openai2023GPT4},
LLaMA2-7B~\citep{touvron2023llama},
LLaMA2-70B~\citep{touvron2023llama},
Vicuna-7B~\citep{chiang2023vicuna}, and
Vicuna-33B~\citep{zheng2023judging}. 
For the task of manual evaluation, we enlist the help of three graduate students specializing in education. We begin by informing the annotators about the purpose of each task and the scoring guidelines. Following this, we request them to score the responses. During the evaluation phase, we randomly select 1000 questions generated by each LLM for each task and ask three volunteers to manually evaluate the generated responses using the same criteria applied to GPT-4.
To ensure the reliability and validity of the human ratings, we calculate the Inter-rater Agreement using Krippendorff’s Alpha (IRA). For ratings that exhibit low agreement ($<$ 0.7), we remove the particular statement from consideration and replace it with a new one. This method guarantees the precision and consistency of our human assessment.

\subsection{Main results}

\emph{Question 1: Which LLM is more suitable to be a general course teacher? Answer 1: GPT4!}
The term ``suitable'' is used to assess the effectiveness of each LLM in different academic subjects. This evaluation helps in identifying which LLMs perform best in specific educational disciplines.
In the general domain tasks, various LLMs demonstrate diverse performances as shown in Table~\ref{tab:xuexi-exp1} and Table~\ref{tab:xuexi-exp4}. Specifically, GPT4 achieves a perfect score of 100\% in both consistency and relevance, indicating its strong capability in understanding task requirements and generating relevant questions. However, its coverage score of 54.5\% suggests there is room for improvement in generating questions that encompass more content. In representativeness, with a score of 80.1\%, GPT-4 shows a good grasp of the context's core content and viewpoints, crafting questions with depth and breadth.
BLOOM-176B and Claude2 also score perfectly in relevance, reflecting their excellent performance in linking questions to the context's themes and content. However, their lower scores in coverage and representativeness indicate potential for improvement in capturing the full extent and core insights of the texts.
Moreover, ``Aver'' in Table~\ref{tab:xuexi-exp1}, Table~\ref{tab:xuexi-exp2} and Table~\ref{tab:xuexi-exp3} represent the average result of the corresponding dimensions under each domain, which are obtained using in-context learning (i.e. ICL). ICL is to introduce a human-written sample into the prompt which typically improves LLMs' performance across all metrics, while most LLMs show a decline in the 0-shot setting, demonstrating the critical role of ICL in enhancing the quality of question generation.

\begin{table}[]
\centering
\resizebox{0.42\textwidth}{!}{%
\begin{tabular}{@{}lcccc|ccc@{}}
\toprule
             & Con & Rel & Cov & Rep & Aver & 0-shot & ↓(\%) \\ \midrule
BLOOM-7B     & 45.5        & 56.4      & 26.6     & 32.1               & 40.2          & 30.8             & 30.4  \\
BLOOM-176B   & \underline{97.7}        & \textbf{100.0}     & \underline{53.7}     & \underline{78.2}               & \underline{82.4}          & \underline{78.3}             & 5.2   \\
Claude2     & 93.7        & \textbf{100.0}     & 40.5     & 68.8               & 75.8          & 74.7             & 1.4   \\
Falcon-7B    & 37.8        & 49.4      & 21.5     & 29.6               & 34.6          & 21.1             & \textbf{63.9}  \\
Falcon-180b & 85.4        & 95.7      & 34.8     & 55.1               & 67.8          & 57.2             & 18.4  \\
GPT3.5       & 96.2        & \textbf{100.0}     & 46.4     & 63.6               & 76.6          & 71.2             & 7.5   \\
GPT4         & \textbf{98.5}        & \textbf{100.0}     & \textbf{54.5}     & \textbf{80.1}               & \textbf{83.3}          & \textbf{80.1}             &4.0    \\
LLaMA2-7B   & 70.2        & 75.2      & 24.7     & 40.0               & 52.5          & 40.7             & 29.1  \\
LLaMA2-70b  & 91.2        & \underline{98.5}      & 39.7     & 65.6               & 73.8          & 71.2             & 3.6   \\
Vicuna-7B    & 68.3        & 70.3      & 31.8     & 42.7               & 53.3          & 38.5             & \underline{38.4}  \\
Vicuna-33B   & 89.6        & 97.2      & 41.7     & 67.3               & 74.0          & 64.6             & 14.5  \\ \bottomrule
\end{tabular}%
}
\caption{Performance of different LLMs in the general domain. Con: Consistency, Rel: Relevance; Cov: Coverage, Rep: Representativeness.}
\label{tab:xuexi-exp1}
\end{table}

\emph{Question 2: Which of the two LLMs is more suitable to act as a humanities teacher and a science teacher? Answer 2: Both are GPT4!}

In the monodisciplinary domain tasks, LLMs are compared based on their performance in humanities and sciences as illustrated in Table~\ref{tab:xuexi-exp2} and Table~\ref{tab:xuexi-exp4}. 
The results reveal that the majority of the LLMs perform marginally better in the scientific disciplines compared to the general domain. 
Specifically, GPT4 excels across all metrics, particularly in the science disciplines, where it scores higher than in the humanities, indicating great capability in handling science content.
Following closely is Claude2, which nearly matches or equals GPT4 in Relevance and representativeness in the humanities, demonstrating a deep understanding and effective processing of humanities content. Claude2 also maintains a high performance in the science disciplines.
GPT3.5 shows competitive strength across the four metrics, especially in relevance and representativeness within the science subjects, approaching the leading performance of GPT4.
BLOOM-176B scores significantly higher in consistency within science compared to the humanities, and also demonstrates good capability in coverage and representativeness, suggesting its strengths in processing logical and scientific data.

\begin{table}[]
\centering
\resizebox{0.48\textwidth}{!}{%
\begin{tabular}{@{}lcc|cc|cc|cc|ccc@{}}
\toprule
            & \multicolumn{2}{c|}{Con} & \multicolumn{2}{c|}{Rel} & \multicolumn{2}{c|}{Cov} & \multicolumn{2}{c|}{Rep} & \multirow{2}{*}{Aver} & \multirow{2}{*}{0-shot} & \multirow{2}{*}{↓(\%)} \\ 
            & H          & S          & H          & S          & H          & S          & H          & S          &                      &                         &                        \\ \midrule
BLOOM-7B    & 28.7       & 29.5       & 61.1       & 62.0         & 47.8       & 51.3       & 55.9       & 54.4       & 48.8                 & 39.3                    & 24.3                   \\
BLOOM-176B  & 58.3       & 71.3       & 88.1       & 92.2       & 62.6       & 76.7       & 71.1       & 72.5       & 74.1                 & 71.5                    & 3.6                    \\
Claude2     & 66.5       & 78.4       & \textbf{95.5}       & \textbf{95.0}         & \underline{75.5}       & \underline{83.4}       & \textbf{82.6}       & 79.7       & 82.1                 & \underline{81.2}                    & 1.1                    \\
Falcon-7B   & 33.2       & 40.1       & 62.3       & 70.1       & 52.3       & 58.8       & 53.5       & 58.4       & 53.6                 & 40.7                    & \underline{31.7}                   \\
Falcon-180B & 53.4       & 70.8       & 88.2       & 91.5       & 71.7       & 75.6       & 67.5       & 70.3       & 73.6                 & 66.8                    & 10.2                   \\
GPT3.5      & \underline{71.9}       & \underline{83.3}       & 92.1       & 93.5       & 74.6       & 79.6       & 75.8       & \underline{79.8}       & \underline{81.3}                 & 79.6                    & 2.2                    \\
GPT4        & \textbf{77.2}       & \textbf{85.4}       & \underline{93.4}       & \underline{94.2}       & \textbf{81.9}       & \textbf{86.2}       & \underline{81.6}       & \textbf{80.3}       & \textbf{85.0}                   & \textbf{84.1}                    & 1.1                    \\
LLaMA2-7B   & 42.1       & 43.9       & 74.5       & 76.3       & 64.6       & 70.2       & 60.2       & 71.8       & 62.9                 & 50.6                    & 24.4                   \\
LLaMA2-70B  & 60.1       & 72.9       & 89         & 89.9       & 73.3       & 78.7       & 73.2       & 76.5       & 76.7                 & 72.3                    & 6.1                    \\
Vicuna-7B   & 42.4       & 48.8       & 72.4       & 76.4       & 56.8       & 62.3       & 60.1       & 65.5       & 60.6                 & 45.5                    & \textbf{33.2}                   \\
Vicuna-33B  & 57.9       & 66.9       & 84.2       & 83.7       & 66.6       & 70.6       & 68.8       & 69.7       & 71.0                   & 65.9                    & 7.8                    \\ \bottomrule
\end{tabular}%
}
\caption{Performance of different LLMs in the monodisciplinary domain.}
\label{tab:xuexi-exp2}
\end{table}

\emph{Question 3: Which LLM is more suitable to be a interdisciplinary teacher? Answer 3: Claude2!}

Results of LLMs in the interdisciplinary domain tasks are shown in Table~\ref{tab:xuexi-exp3} and Table~\ref{tab:xuexi-exp4}. It shows that Claude2 outperforms other LLMs with scores of 89.1\% in relevance and 93.3\% in representativeness. Following closely is GPT4, with scores of 87.8\% in relevance and 91.2\% in representativeness, also indicating strong performance.
GPT3.5 and LLaMA2-70B also show high scores, particularly in representativeness, suggesting their capability in understanding key textual content and generating in-depth questions.
On the other hand, BLOOM-7B, Falcon-7B, and Vicuna-7B perform relatively poorly on both metrics. Specifically, BLOOM-7B scores below 40\% in both relevance and representativeness, which may suggest a need for further enhancement in understanding interdisciplinary content and generating high-quality questions.

\begin{table}[]
\centering
\resizebox{0.3\textwidth}{!}{%
\begin{tabular}{@{}lcc|ccc@{}}
\toprule
            & Rel  & Rep  & Aver  & 0-shot & ↓(\%) \\\midrule
BLOOM-7B    & 29.9 & 36.2 & 33.1 & 22.5   & 46.9  \\
BLOOM-176B  & 77.2 & 80.2 & 78.7 & 68.5   & 14.9  \\
Claude2     & \textbf{89.1} & \textbf{93.3} & \textbf{91.2} & \textbf{88.8}   & 2.7   \\
Falcon-7B   & 27.8 & 33.6 & 30.7 & 20.6   & \textbf{49.0}    \\
Falcon-180B & 79.8 & 80.0   & 79.9 & 71.7   & 11.4  \\
GPT3.5      & 85.6 & 90.1 & 87.9 & 80.7   & 8.9   \\
GPT4        & \underline{87.8} & \underline{91.2} & \underline{89.5} & \underline{86.3}   & 3.7   \\
LLaMA2-7B   & 50.5 & 58.7 & 54.6 & 39.8   & 37.2  \\
LLaMA2-70B  & 82.3 & 88.9 & 85.6 & 81.6   & 4.9   \\
Vicuna-7B   & 47.2 & 49.8 & 48.5 & 32.7   & \underline{48.3}  \\
Vicuna-33B  & 72.4 & 80.2 & 76.3 & 68.6   & 11.2 \\
\bottomrule
\end{tabular}%
}
\caption{Performance of different LLMs in the interdisciplinary domain.}
\label{tab:xuexi-exp3}
\end{table}

\emph{Question 4: Which LLM is more suitable to be a all-around teacher? Answer 4: GPT4!}

\begin{figure*}[!h]
  \centering
  \includegraphics[width=\linewidth]{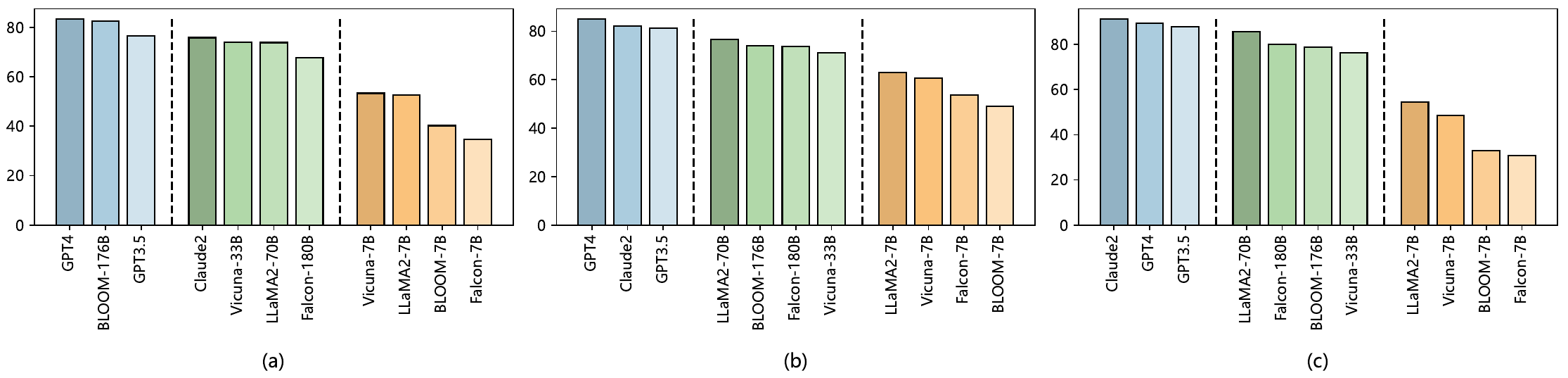}
  \caption{The comprehensive performance of different LLMs in the proposed Dr.Academy benchmark.}
  \label{fig:xuexi-exp1}
\end{figure*}

We also comprehensively compare the performance of LLMs in three tasks as shown in Table~\ref{tab:xuexi-exp4} and Fig.~\ref{fig:xuexi-exp1}.
Specifically, in the general domain tasks, GPT4 scores the highest and ranks first. In the monodisciplinary domain tasks, including humanities and science, GPT4 also has the best performance. In the interdisciplinary domain tasks, Claude2 occupies the first rank. Finally, for the comprehensive rating, GPT4 ranks first again with the highest score. Overall, GPT4 shows the best performance in most tasks, while Claude2 also demonstrates strong capability in certain tasks.
Overall, GPT4, Claude2, and GPT3.5 perform well in these assessments, demonstrating their versatility and adaptability as high-performance models. On the other hand, BLOOM-7B and Falcon-7B tend to perform weaker in most fields, which may make them more suitable for specific application scenarios. 

\begin{table}[]
\centering
\resizebox{0.47\textwidth}{!}{%
\begin{tabular}{@{}lcc|cc|cc|cc|cc|cc@{}}
\toprule
            & \multicolumn{2}{c|}{Gen} & \multicolumn{2}{c|}{Mon-H} & \multicolumn{2}{c|}{Mon-S} & \multicolumn{2}{c|}{Mon} & \multicolumn{2}{c|}{Int} & \multicolumn{2}{c}{Com} \\ 
            & S                & R    & S                 & R     & S                 & R     & S                & R    & S                & R    & S                & R    \\\midrule
GPT4        & \textbf{83.3}    & 1    & \textbf{83.5}     & 1     & \textbf{86.5}     & 1     & \textbf{85.0}    & 1    & \underline{89.5}    & 2    & \textbf{85.9}    & 1    \\
Claude2     & 75.8            & 4    & \underline{80.0}              & 2     & \underline{84.1}              & 2     & \underline{82.1}             & 2    & \textbf{91.2}             & 1    & \underline{83.0}             & 2    \\
GPT3.5      & 76.6             & 3    & 78.6              & 3     & 84.1              & 3     & 81.3             & 3    & 87.9             & 3    & 81.9             & 3    \\
LLaMA2-70B  & 73.8             & 6    & 73.9              & 4     & 79.5              & 4     & 76.7             & 4    & 85.6             & 4    & 78.7             & 4    \\
BLOOM-176B  & \underline{82.4}             & 2    & 70.0              & 6     & 78.2              & 5     & 74.1             & 5    & 78.7             & 6    & 78.4             & 5    \\
Falcon-180B & 67.8             & 7    & 70.2              & 5     & 77.1              & 6     & 73.7             & 6    & 79.9             & 5    & 73.8             & 6    \\
Vicuna-33B  & 74.0             & 5    & 69.4              & 7     & 72.7              & 7     & 71.1             & 7    & 76.3             & 7    & 73.8             & 7    \\
LLaMA2-7B   & 52.5             & 9    & 60.3              & 8     & 65.6              & 8     & 63.0             & 8    & 54.6             & 8    & 56.7             & 8    \\
Vicuna-7B   & 53.3             & 8    & 57.9              & 9     & 63.3              & 9     & 60.6             & 9    & 48.5             & 9    & 54.1             & 9    \\
BLOOM-7B    & 40.2             & 10   & 48.4              & 11    & 49.3              & 11    & 48.9             & 11   & 33.1             & 10   & 40.7             & 10   \\
Falcon-7B   & 34.6             & 11   & 50.3              & 10    & 56.9              & 10    & 53.6             & 10   & 30.7             & 11   & 39.6             & 11   \\ \bottomrule
\end{tabular}%
}
\caption{Score and rank of different LLMs among general, humanities-related monodisciplinary, science-related monodisciplinary, and interdisciplinary domain. And LLMs' comprehensive performance. S: score, R: rank, Gen: the general domain, Mon-H: humanities-related monodisciplinary domain, Mon-S: science-related monodisciplinary domain, Int: the interdisciplinary domain, Com: the comprehensive performance.}
\label{tab:xuexi-exp4}
\end{table}

\emph{Question 5: What's the relationship among metrics for various LLMs? Answer 5: Pairwise positive correlations!}

\begin{figure}[!h]
  \centering
  \includegraphics[width=\linewidth]{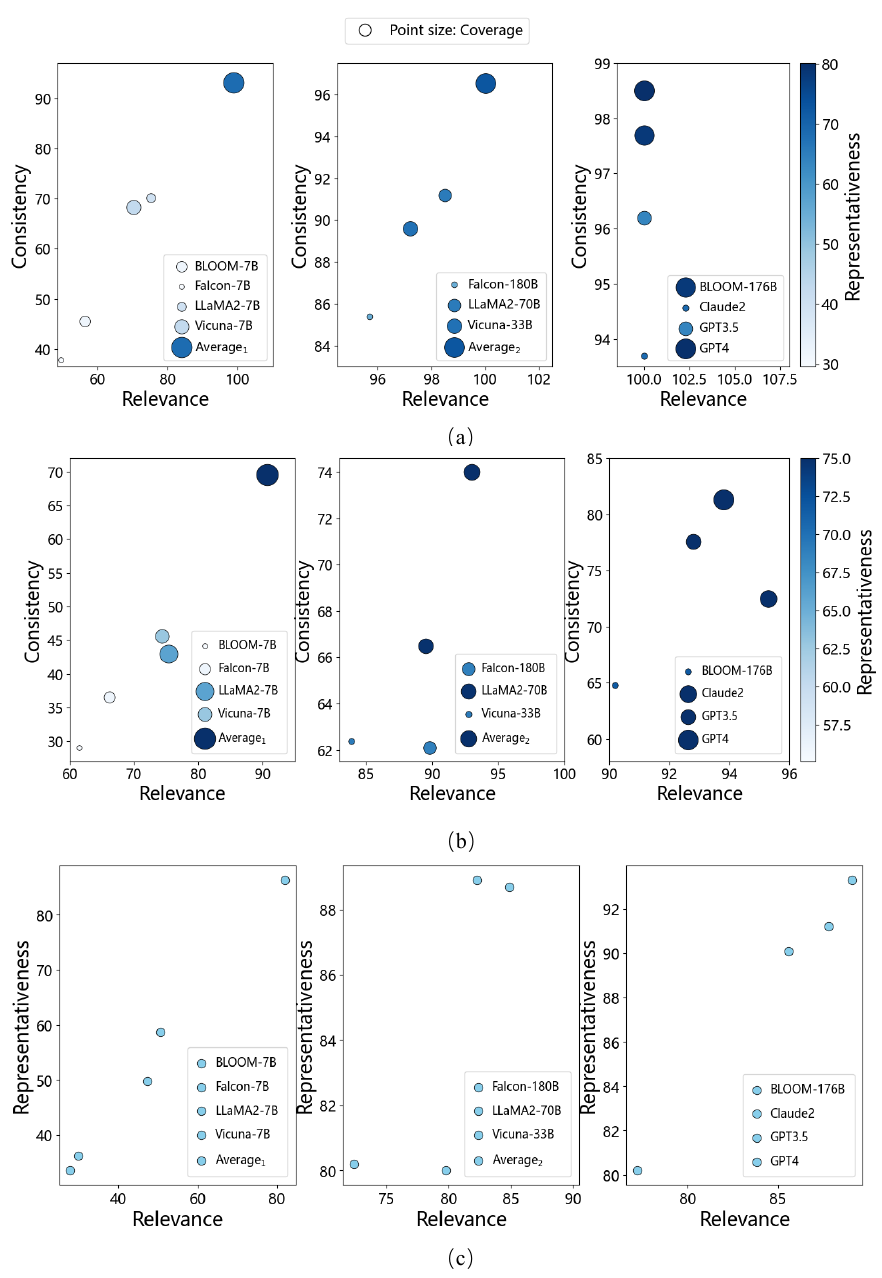}
  \caption{The relationship among four metrics of different LLMs.}
  \label{fig:xuexi-exp2}
\end{figure}

We also analyze the relationship among four metrics in three tasks as shown in Fig.~\ref{fig:xuexi-exp2}. 
Fig.~\ref{fig:xuexi-exp2} (a) and Fig.~\ref{fig:xuexi-exp2} (b) represent the relationship between four metrics of question quality generated by different LLMs in the general and monodisciplinary domains, respectively. The size of the circle indicates coverage, with larger circles covering more content of the text. The darker the color of the circle, the higher the relevance of the questions to the text. The ``Average$_1$'' in the first graph represents a group of LLMs, which are zoomed in on the second graph, and the ``Average$_2$'' in the second graph represents a subset of these LLMs, which are further examined in the third graph.
In Fig.~\ref{fig:xuexi-exp2} (a), analyzing the general domain tasks, we see a positive correlation between relevance and consistency across all three graphs. Representativeness also shows a positive correlation with relevance and consistency, but the correlation is weaker. As relevance and consistency increase, the LLMs have darker colors and larger circles, indicating higher relevance and coverage. Although larger LLMs like BLOOM-176B show good coverage, not all models with large coverage have high relevance. For example, Falcon-180B does not perform as well as BLOOM-176B in relevance, suggesting a need for balance between the breadth of text coverage and the accuracy of question generation. In the third graph, LLMs like GPT4 maintain high relevance while also achieving good coverage.
In Fig.~\ref{fig:xuexi-exp2} (b), for the monodisciplinary domain tasks, the correlations between the four metrics are not as pronounced as in the general domain. The third graph shows little color variation, indicating that representativeness does not change much with increased relevance and consistency. However, there are LLMs like GPT4 that stand out in all metrics, shown in the top right corner with a dark color and large size. But GPT3.5, while showing good representativeness and relevance, has only average consistency.
In Fig.~\ref{fig:xuexi-exp2} (c), analyzing the interdisciplinary domain tasks, generally shows a positive correlation between relevance and representativeness, although it's not as clear in the second graph.
Overall, LLMs like GPT4, Claude2, BLOOM-176B, and GPT3.5 perform well across all four metrics, while the 7B series models tend to perform less well. The metrics also tend to show positive correlations with each other.

\emph{Question 6: Is the automatic scores generated by GPT4 agree with human perspectives? Answer 6: Yes, the Pearson correlation coefficient reaches 0.947 and Spearman rank correlation reaches 0.87!}

\begin{figure}[!h]
  \centering
  \includegraphics[width=\linewidth]{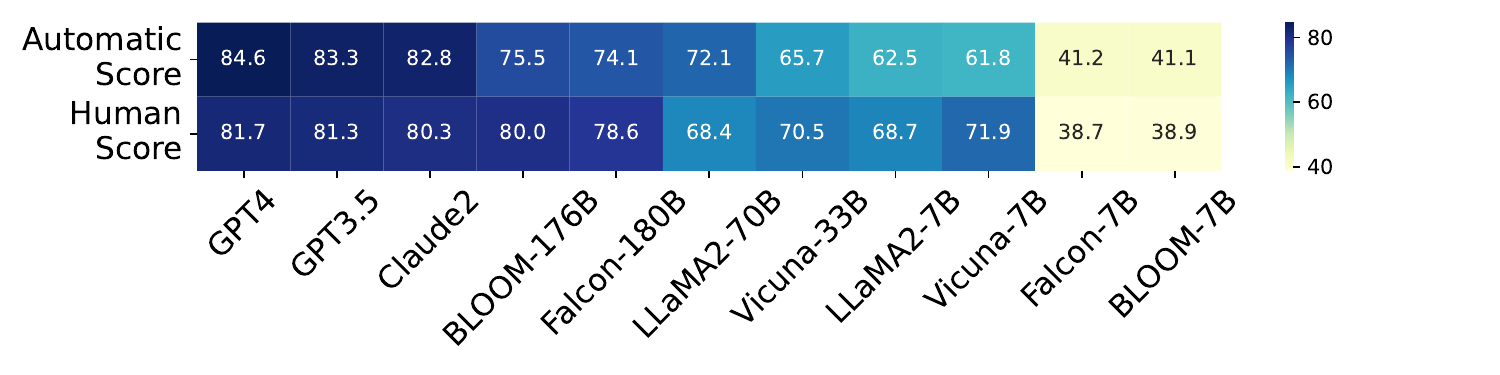}
  \caption{The Pearson correlation coefficient that normalized to a 1-100 scale between automatic and human scores of different LLMs.}
  \label{fig:xuexi-exp3}
\end{figure}

We adopt Pearson correlation coefficient that normalized to a 1-100 scale to investigate the difference between automatic and human scores for different LLMs as shown in Fig.~\ref{fig:xuexi-exp3}. 
The human scores for each metric and the corresponding agreement of human annotators on these metrics are listed in Table~\ref{tab:agreement}.
We find that automatic and human evaluations for LLMs generally agree, showing a high positive correlation with the Pearson correlation coefficient reaching 0.947 and Spearman rank correlation reaching 0.870.
Specifically, GPT4 performs excellently in both automatic and human scoring, with minimal difference, indicating widespread recognition of its capability. Similarly, Claude2 has close scores in both evaluations, indicating balanced performance in assessment tasks.
It's important to clarify that the process of generating questions and scoring them is separate. During the scoring phase, there is no knowledge of which LLM generates which question. We believe that even if other LLMs are used to evaluate GPT-4's performance against a comparatively weaker 7b model, the results would still favor GPT-4, a conclusion also supported by human evaluations. 
The findings suggests that automatic scoring has the potential to partially replace human scoring for evaluating questioning capability in education as a teacher of LLMs through evaluating their generated educational questions.

\begin{table}[]
\centering
\resizebox{0.47\textwidth}{!}{%
\begin{tabular}{@{}lcc|cc|cc|cc@{}}
\toprule
            & \multicolumn{2}{c|}{Consistency} & \multicolumn{2}{c|}{Relevance} & \multicolumn{2}{c|}{Coverage} & \multicolumn{2}{c}{Representativeness} \\ 
            & Score        & Agreement        & Score       & Agreement       & Score       & Agreement      & Score            & Agreement           \\\midrule
BLOOM-7B    & 30.8         & 0.87             & 38.5        & 0.91            & 46.8        & 0.82           & 39.5             & 0.75                \\
BLOOM-176B  & 81.5         & 0.92             & 86.2        & 0.93            & 77.1        & 0.79           & 75.2             & 0.77                \\
Claude2    & 78.6         & 0.95             & 88.8        & 0.95            & 76.7        & 0.83           & 77.1             & 0.82                \\
Falcon-7B   & 31.5         & 0.88             & 42.8        & 0.9             & 39.8        & 0.78           & 40.7             & 0.74                \\
Falcon-180B & 77.4         & 0.90              & 85.6        & 0.91            & 77.4        & 0.8            & 74               & 0.78                \\
GPT3.5      & 78.6         & 0.94             & 86.3        & 0.92            & 77.9        & 0.82           & 71.6             & 0.81                \\
GPT4        & 83.8         & 0.95             & 88.3        & 0.94            & 78.3        & 0.83           & 76.4             & 0.83                \\
LLaMA 2-7B  & 60.8         & 0.85             & 73.4        & 0.86            & 70.7        & 0.76           & 69.9             & 0.8                 \\
LLaMA 2-70B & 63.5         & 0.89             & 74.7        & 0.92            & 66.6        & 0.79           & 68.8             & 0.81                \\
Vicuna-7B   & 69.9         & 0.84             & 78.4        & 0.87            & 69          & 0.77           & 70.3             & 0.76                \\
Vicuna-33B  & 70.2         & 0.88             & 76.5        & 0.89            & 64.6        & 0.81           & 70.7             & 0.78                \\ \bottomrule
\end{tabular}%
}
\caption{The human scores for each metric and the corresponding agreement of human annotators on these metrics.}
\label{tab:agreement}
\end{table}


\begin{figure}[!h]
  \centering
  \includegraphics[width=\linewidth]{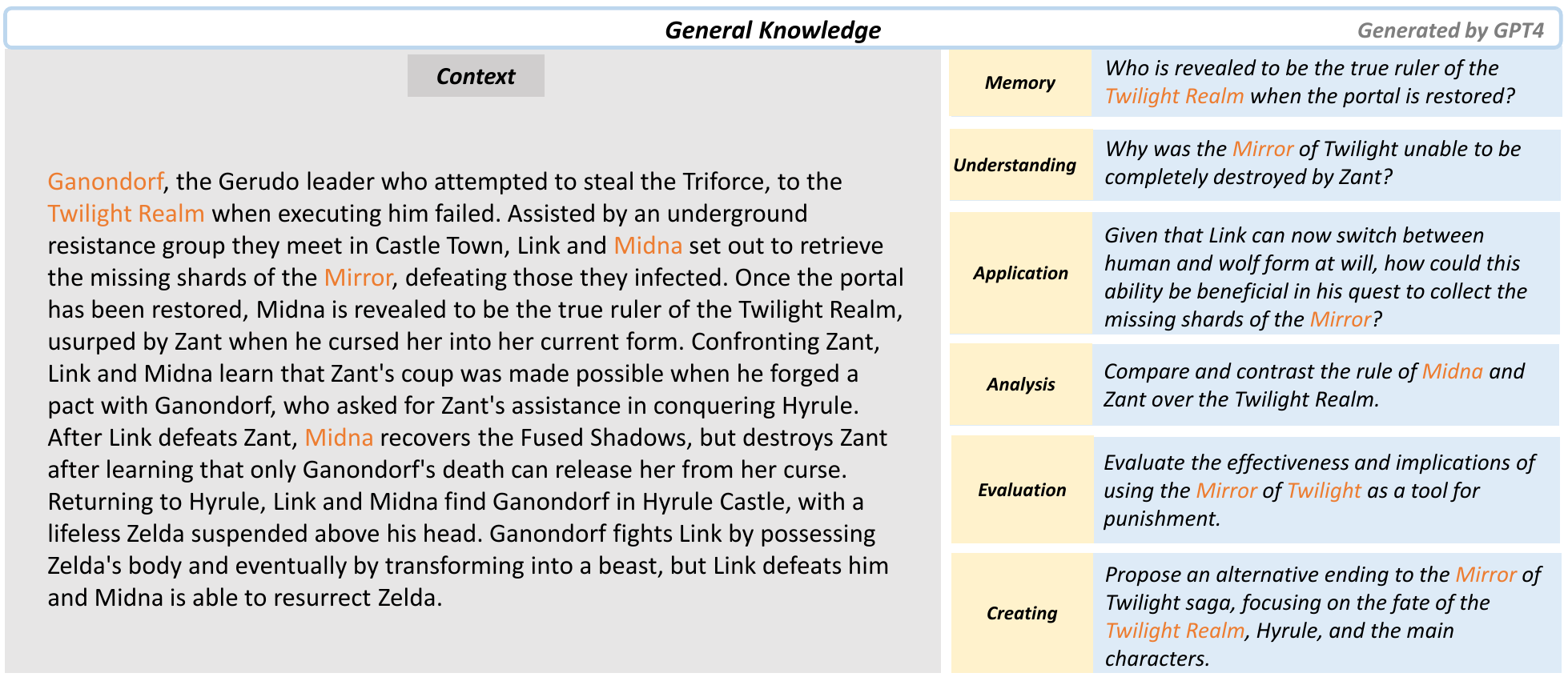}
  \caption{Questions posed by the top LLM, i.e. GPT4, in the general domain.}
  \label{fig:xuexi-case1}
\end{figure}

\begin{figure}[!h]
  \centering
  \includegraphics[width=\linewidth]{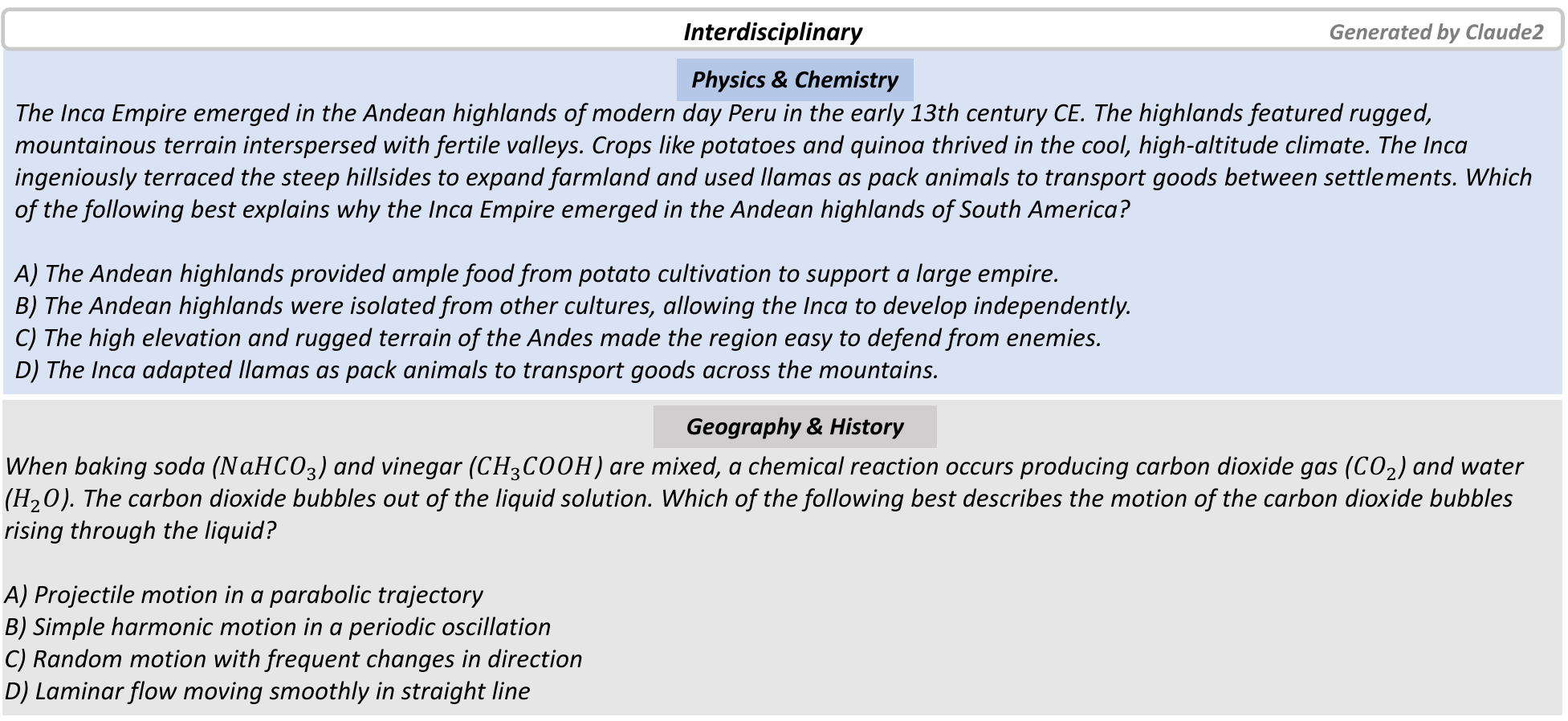}
  \caption{Questions posed by the top LLM, i.e. Claude2, in the interdisciplinary domain.}
  \label{fig:xuexi-case4}
\end{figure}

\subsection{Case study}

We present a set of general domain questions generated by GPT4, identified as the leading LLM in this area, in Fig.\ref{fig:xuexi-case1}. The top-performing LLMs in monodisciplinary (Humanities), monodisciplinary (Science), and interdisciplinary domains have their questions displayed in Fig.\ref{fig:xuexi-case2}, Fig.\ref{fig:xuexi-case3}, and Fig.\ref{fig:xuexi-case4}, respectively. Additionally, further examples from baseline LLMs are shown in figures ranging from Fig.\ref{fig:1xuexi-case1} to Fig.\ref{fig:4xuexi-case6}.
In Fig.\ref{fig:xuexi-case1}, the first question tests memory by asking about the true ruler of the Twilight Realm. The second requires understanding the reasons behind the Mirror of Twilight's durability. The third applies Link's transformation ability to his quest. The fourth analyzes the contrasting rules of Midna and Zant. The fifth evaluates the use of the Mirror as a punishment tool. Finally, the sixth encourages creating an alternative ending for the saga. These questions showcase GPT-4's range from simple recall to creative thinking, and the generated questions can be adopted to make students quickly grasp the key content, indicating its strong questioning capability in education as a teacher through evaluating their generated educational questions.
We also show a good case in the interdisciplinary tasks generated by Claude2, which shows outstanding comprehensive capability that can effectively combining different subject knowledge to pose accurate and challenging questions. In chemistry and physics combination questions, it understands the basic principles of chemical reactions and the physical properties of bubble motion in liquids, demonstrating analysis and synthesis ability in interdisciplinary questions. Claude2 shows high adaptability and understanding in complex tasks involving multiple disciplines.

\section{Related Work}

\subsection{Question Generation}
The potential of LLMs in generating questions for educational purposes garners significant academic attention. \citet{chen2019reinforcement} have developed a reinforcement learning approach specifically for generating natural questions. \citet{elkins2023useful} evaluate the educational utility of questions produced by LLMs, organizing them according to their levels of difficulty. \citet{tavares2023learning} investigate the methods LLMs use to create questions, focusing on the tracking of dialogue states. \citet{kai2021learning} introduce a method involving double hints for the generation of questions about visuals. \citet{uehara2022learning} emphasize the importance of sub-questions in improving the effectiveness of primary visual questions. \citet{arora2022ask} examine various prompting techniques for LLMs and analyze the differences in model responses. \citet{abdelghani2022gpt} utilize the capabilities of GPT-3 to foster curiosity-driven questioning among children. Collectively, these studies underline the evolving role of LLMs in reshaping question generation instead of searching valuable questioning points.

\subsection{Test-based Benchmark}
There has been an increasing focus on evaluating the capability of LLMs in the context of standardized exams and academic benchmarks.
For example, 
~\citet{zhang2023evaluating} introduce GAOKAO Benchmark to evaluate the intuitive benchmark of Chinese college entrance examination questions;
~\citet{huang2023c} propose the first comprehensive Chinese evaluation package C-EVAL;
~\citet{zhong2023agieval} present a human-centric benchmark AGIEval designed for evaluating foundation models;
~\citet{li2023cmmlu} introduce CMMLU, a comprehensive Chinese benchmark covering multiple disciplines;
~\citet{wang2023scibench} introduce SciBench to systematically investigate the reasoning ability required to solve complex scientific problems;;
~\citet{liu2023m3ke} propose M3KE, a large-scale multi-layer and multi-disciplinary knowledge assessment benchmark;
~\citet{zeng2023measuring} propose a method to evaluate the multi-task accuracy of large Chinese language models across various domains;
~\citet{zhang2023fineval} introduce FinEval, a benchmark designed for financial knowledge evaluation in LLMs;
~\citet{wei2023cmath} focus on the Chinese Elementary School Math Word Problems dataset to evaluate reasoning capability;
~\citet{dao2023can} explore ChatGPT's potential to complete the Vietnam National High School Graduation Exam;
\citet{raina2022multiple} propose performance criteria to assess the generated multiple-choice questions;
\citet{chen2018learningq} present LearningQ, an educational question generation dataset containing over 230K document-question pairs.
\citet{chen2024dolarge} introduce a novel game named BrainKing for evaluating LLM capabilities under incomplete information scenarios.
\citet{chen2024emotionqueen} presents a novel framework named EmotionQueen for evaluating the emotional intelligence of LLMs. 
However, these tests treat LLMs as students, assessing their abilities by how they answer questions instead of seeing them as teachers.

\section{Conclusions and Future Work}
In conclusion, our study presents a pioneering investigation into the questioning capability in education as a teacher of large language models (LLMs) through evaluating their generated educational questions, shifting the traditional role of LLMs from learners to educators. We have developed a comprehensive benchmark, named Dr.Academy, based on educational taxonomies that assesses LLMs' abilities to generate questions across various domains with four evaluation metrics. Our findings indicate that models like GPT4, Claude2, and GPT3.5 demonstrate promising teaching potential. Looking ahead, the future directions of this research include refining the evaluation metrics for even more nuanced assessments of teaching effectiveness and expanding the range of subjects and domains covered.

\section*{Limitations}

One limitation of our study is that it primarily focuses on the ability of large language models (LLMs) to generate questions, which is just one aspect of teaching. Actual teaching involves more complex interactions, including providing feedback, adapting to students' needs, and fostering critical thinking, areas not fully captured by our current benchmark. Additionally, our approach relies heavily on the textual content, which may not comprehensively represent the nuances of human teaching methods that include non-verbal cues and personalized interactions. Therefore, while our findings offer valuable insights into the potential of LLMs as teaching aids, they should be viewed as a starting point for more in-depth research into the multifaceted nature of teaching and learning processes.

\section*{Acknowledgements }
This work is supported by
Science and Technology Commission of Shanghai Municipality Grant (No. 22511105902), Shanghai Municipal Science and Technology Major Project (No.2021SHZDZX0103), the National Natural Science Foundation of China (No.62072323), Shanghai Science and Technology Innovation Action Plan (No. 22511104700), and the Zhejiang Lab Open Research Project (NO. K2022NB0AB04).

\bibliography{anthology,acl}

\begin{thebibliography}{54}
\expandafter\ifx\csname natexlab\endcsname\relax\def\natexlab#1{#1}\fi

\bibitem[{Abdelghani et~al.(2022)Abdelghani, Wang, Yuan, Wang, Sauz{\'e}on, and Oudeyer}]{abdelghani2022gpt}
Rania Abdelghani, Yen-Hsiang Wang, Xingdi Yuan, Tong Wang, H{\'e}l{\`e}ne Sauz{\'e}on, and Pierre-Yves Oudeyer. 2022.
\newblock Gpt-3-driven pedagogical agents for training children's curious question-asking skills.
\newblock \emph{arXiv preprint arXiv:2211.14228}.

\bibitem[{Almazrouei et~al.(2023)Almazrouei, Alobeidli, Alshamsi, Cappelli, Cojocaru, Debbah, Étienne Goffinet, Hesslow, Launay, Malartic, Mazzotta, Noune, Pannier, and Penedo}]{almazrouei2023falcon}
Ebtesam Almazrouei, Hamza Alobeidli, Abdulaziz Alshamsi, Alessandro Cappelli, Ruxandra Cojocaru, Mérouane Debbah, Étienne Goffinet, Daniel Hesslow, Julien Launay, Quentin Malartic, Daniele Mazzotta, Badreddine Noune, Baptiste Pannier, and Guilherme Penedo. 2023.
\newblock \href {http://arxiv.org/abs/2311.16867} {The falcon series of open language models}.

\bibitem[{Anderson and Krathwohl(2001)}]{anderson2001taxonomy}
Lorin~W Anderson and David~R Krathwohl. 2001.
\newblock \emph{A taxonomy for learning, teaching, and assessing: A revision of Bloom's taxonomy of educational objectives: complete edition}.
\newblock Addison Wesley Longman, Inc.

\bibitem[{Arora et~al.(2022)Arora, Narayan, Chen, Orr, Guha, Bhatia, Chami, Sala, and R{\'e}}]{arora2022ask}
Simran Arora, Avanika Narayan, Mayee~F Chen, Laurel~J Orr, Neel Guha, Kush Bhatia, Ines Chami, Frederic Sala, and Christopher R{\'e}. 2022.
\newblock Ask me anything: A simple strategy for prompting language models.
\newblock \emph{arXiv preprint arXiv:2210.02441}.

\bibitem[{Bai et~al.(2022)Bai, Kadavath, Kundu, Askell, Kernion, Jones, Chen, Goldie, Mirhoseini, McKinnon, Chen, Olsson, Olah, Hernandez, Drain, Ganguli, Li, Tran-Johnson, Perez, Kerr, Mueller, Ladish, Landau, Ndousse, Lukosuite, Lovitt, Sellitto, Elhage, Schiefer, Mercado, DasSarma, Lasenby, Larson, Ringer, Johnston, Kravec, Showk, Fort, Lanham, Telleen-Lawton, Conerly, Henighan, Hume, Bowman, Hatfield-Dodds, Mann, Amodei, Joseph, McCandlish, Brown, and Kaplan}]{bai2022constitutional}
Yuntao Bai, Saurav Kadavath, Sandipan Kundu, Amanda Askell, Jackson Kernion, Andy Jones, Anna Chen, Anna Goldie, Azalia Mirhoseini, Cameron McKinnon, Carol Chen, Catherine Olsson, Christopher Olah, Danny Hernandez, Dawn Drain, Deep Ganguli, Dustin Li, Eli Tran-Johnson, Ethan Perez, Jamie Kerr, Jared Mueller, Jeffrey Ladish, Joshua Landau, Kamal Ndousse, Kamile Lukosuite, Liane Lovitt, Michael Sellitto, Nelson Elhage, Nicholas Schiefer, Noemi Mercado, Nova DasSarma, Robert Lasenby, Robin Larson, Sam Ringer, Scott Johnston, Shauna Kravec, Sheer~El Showk, Stanislav Fort, Tamera Lanham, Timothy Telleen-Lawton, Tom Conerly, Tom Henighan, Tristan Hume, Samuel~R. Bowman, Zac Hatfield-Dodds, Ben Mann, Dario Amodei, Nicholas Joseph, Sam McCandlish, Tom Brown, and Jared Kaplan. 2022.
\newblock \href {http://arxiv.org/abs/2212.08073} {Constitutional ai: Harmlessness from ai feedback}.

\bibitem[{Bai et~al.(2023)Bai, Lv, Zhang, Lyu, Tang, Huang, Du, Liu, Zeng, Hou et~al.}]{bai2023longbench}
Yushi Bai, Xin Lv, Jiajie Zhang, Hongchang Lyu, Jiankai Tang, Zhidian Huang, Zhengxiao Du, Xiao Liu, Aohan Zeng, Lei Hou, et~al. 2023.
\newblock Longbench: A bilingual, multitask benchmark for long context understanding.
\newblock \emph{arXiv preprint arXiv:2308.14508}.

\bibitem[{Brown et~al.(2020)Brown, Mann, Ryder, Subbiah, Kaplan, Dhariwal, Neelakantan, Shyam, Sastry, Askell, Agarwal, Herbert-Voss, Krueger, Henighan, Child, Ramesh, Ziegler, Wu, Winter, Hesse, Chen, Sigler, Litwin, Gray, Chess, Clark, Berner, McCandlish, Radford, Sutskever, and Amodei}]{brown2020language}
Tom~B. Brown, Benjamin Mann, Nick Ryder, Melanie Subbiah, Jared Kaplan, Prafulla Dhariwal, Arvind Neelakantan, Pranav Shyam, Girish Sastry, Amanda Askell, Sandhini Agarwal, Ariel Herbert-Voss, Gretchen Krueger, Tom Henighan, Rewon Child, Aditya Ramesh, Daniel~M. Ziegler, Jeffrey Wu, Clemens Winter, Christopher Hesse, Mark Chen, Eric Sigler, Mateusz Litwin, Scott Gray, Benjamin Chess, Jack Clark, Christopher Berner, Sam McCandlish, Alec Radford, Ilya Sutskever, and Dario Amodei. 2020.
\newblock \href {http://arxiv.org/abs/2005.14165} {Language models are few-shot learners}.

\bibitem[{Chen et~al.(2018)Chen, Yang, Hauff, and Houben}]{chen2018learningq}
Guanliang Chen, Jie Yang, Claudia Hauff, and Geert-Jan Houben. 2018.
\newblock Learningq: a large-scale dataset for educational question generation.
\newblock In \emph{Proceedings of the international AAAI conference on web and social media}, volume~12.

\bibitem[{Chen et~al.(2023{\natexlab{a}})Chen, Zhang, and Zhao}]{chen2023modeling}
Jialin Chen, Zhuosheng Zhang, and Hai Zhao. 2023{\natexlab{a}}.
\newblock Modeling hierarchical reasoning chains by linking discourse units and key phrases for reading comprehension.
\newblock \emph{arXiv preprint arXiv:2306.12069}.

\bibitem[{Chen et~al.(2019)Chen, Wu, and Zaki}]{chen2019reinforcement}
Yu~Chen, Lingfei Wu, and Mohammed~J Zaki. 2019.
\newblock Reinforcement learning based graph-to-sequence model for natural question generation.
\newblock \emph{arXiv preprint arXiv:1908.04942}.

\bibitem[{Chen et~al.(2023{\natexlab{b}})Chen, Fu, Yuan, Wen, Fan, Liu, Zhang, Li, and Xiao}]{chen2023hallucination}
Yuyan Chen, Qiang Fu, Yichen Yuan, Zhihao Wen, Ge~Fan, Dayiheng Liu, Dongmei Zhang, Zhixu Li, and Yanghua Xiao. 2023{\natexlab{b}}.
\newblock Hallucination detection: Robustly discerning reliable answers in large language models.
\newblock In \emph{Proceedings of the 32nd ACM International Conference on Information and Knowledge Management}, pages 245--255.

\bibitem[{Chen et~al.(2024{\natexlab{a}})Chen, Li, Yan, Liu, Liang, and Xiao}]{chen2024dolarge}
Yuyan Chen, Yueze Li, Songzhou Yan, Sijia Liu, Jiaqing Liang, and Yanghua Xiao. 2024{\natexlab{a}}.
\newblock Do large language models have problem-solving capability under incomplete information scenarios?
\newblock In \emph{Proceedings of the 62nd Annual Meeting of the Association for Computational Linguistics}.

\bibitem[{Chen et~al.(2023{\natexlab{c}})Chen, Li, Liang, Xiao, Liu, and Chen}]{chen2023can}
Yuyan Chen, Zhixu Li, Jiaqing Liang, Yanghua Xiao, Bang Liu, and Yunwen Chen. 2023{\natexlab{c}}.
\newblock Can pre-trained language models understand chinese humor?
\newblock In \emph{Proceedings of the Sixteenth ACM International Conference on Web Search and Data Mining}, pages 465--480.

\bibitem[{Chen et~al.(2023{\natexlab{d}})Chen, Wen, Fan, Chen, Wu, Liu, Li, Liu, and Xiao}]{chen2023mapo}
Yuyan Chen, Zhihao Wen, Ge~Fan, Zhengyu Chen, Wei Wu, Dayiheng Liu, Zhixu Li, Bang Liu, and Yanghua Xiao. 2023{\natexlab{d}}.
\newblock Mapo: Boosting large language model performance with model-adaptive prompt optimization.
\newblock In \emph{Findings of the Association for Computational Linguistics: EMNLP 2023}, pages 3279--3304.

\bibitem[{Chen et~al.(2023{\natexlab{e}})Chen, Xiao, Li, and Liu}]{chen2023xmqas}
Yuyan Chen, Yanghua Xiao, Zhixu Li, and Bang Liu. 2023{\natexlab{e}}.
\newblock Xmqas: Constructing complex-modified question-answering dataset for robust question understanding.
\newblock \emph{IEEE Transactions on Knowledge and Data Engineering}.

\bibitem[{Chen et~al.(2022)Chen, Xiao, and Liu}]{chen2022grow}
Yuyan Chen, Yanghua Xiao, and Bang Liu. 2022.
\newblock Grow-and-clip: Informative-yet-concise evidence distillation for answer explanation.
\newblock In \emph{2022 IEEE 38th International Conference on Data Engineering (ICDE)}, pages 741--754. IEEE.

\bibitem[{Chen et~al.(2024{\natexlab{b}})Chen, Yan, Guo, Jia, Li, and Xiao}]{chen2024hotvcom}
Yuyan Chen, Songzhou Yan, Qingpei Guo, Jiyuan Jia, Zhixu Li, and Yanghua Xiao. 2024{\natexlab{b}}.
\newblock Hotvcom: Generating buzzworthy comments for videos.
\newblock In \emph{Proceedings of the 62nd Annual Meeting of the Association for Computational Linguistics}.

\bibitem[{Chen et~al.(2024{\natexlab{c}})Chen, Yan, Liu, Li, and Xiao}]{chen2024emotionqueen}
Yuyan Chen, Songzhou Yan, Sijia Liu, Yueze Li, and Yanghua Xiao. 2024{\natexlab{c}}.
\newblock Emotionqueen: A benchmark for evaluating empathy of large language models.
\newblock In \emph{Proceedings of the 62nd Annual Meeting of the Association for Computational Linguistics}.

\bibitem[{Chen et~al.(2024{\natexlab{d}})Chen, Yan, Zhu, Li, and Xiao}]{chen2024xmecap}
Yuyan Chen, Songzhou Yan, Zhihong Zhu, Zhixu Li, and Yanghua Xiao. 2024{\natexlab{d}}.
\newblock Xmecap: Meme caption generation with sub-image adaptability.
\newblock In \emph{Proceedings of the 32nd ACM Multimedia}.

\bibitem[{Chen et~al.(2024{\natexlab{e}})Chen, Yuan, Liu, Liu, Guan, Guo, Peng, Liu, Li, and Xiao}]{chen2024talk}
Yuyan Chen, Yichen Yuan, Panjun Liu, Dayiheng Liu, Qinghao Guan, Mengfei Guo, Haiming Peng, Bang Liu, Zhixu Li, and Yanghua Xiao. 2024{\natexlab{e}}.
\newblock Talk funny! a large-scale humor response dataset with chain-of-humor interpretation.
\newblock In \emph{Proceedings of the AAAI Conference on Artificial Intelligence}, volume~38, pages 17826--17834.

\bibitem[{Chen et~al.(2024{\natexlab{f}})Chen, Zhao, Wen, Li, and Xiao}]{chen2024temporalmed}
Yuyan Chen, Jin Zhao, Zhihao Wen, Zhixu Li, and Yanghua Xiao. 2024{\natexlab{f}}.
\newblock Temporalmed: Advancing medical dialogues with time-aware responses in large language models.
\newblock In \emph{Proceedings of the 17th ACM International Conference on Web Search and Data Mining}, pages 116--124.

\bibitem[{Cheng et~al.(2023)Cheng, Huang, and Wei}]{cheng2023adapting}
Daixuan Cheng, Shaohan Huang, and Furu Wei. 2023.
\newblock Adapting large language models via reading comprehension.
\newblock \emph{arXiv preprint arXiv:2309.09530}.

\bibitem[{Chiang et~al.(2023)Chiang, Li, Lin, Sheng, Wu, Zhang, Zheng, Zhuang, Zhuang, Gonzalez et~al.}]{chiang2023vicuna}
Wei-Lin Chiang, Zhuohan Li, Zi~Lin, Ying Sheng, Zhanghao Wu, Hao Zhang, Lianmin Zheng, Siyuan Zhuang, Yonghao Zhuang, Joseph~E Gonzalez, et~al. 2023.
\newblock Vicuna: An open-source chatbot impressing gpt-4 with 90\%* chatgpt quality.
\newblock \emph{See https://vicuna. lmsys. org (accessed 14 April 2023)}.

\bibitem[{Chung et~al.(2023)Chung, Kamar, and Amershi}]{chung2023increasing}
John Joon~Young Chung, Ece Kamar, and Saleema Amershi. 2023.
\newblock Increasing diversity while maintaining accuracy: Text data generation with large language models and human interventions.
\newblock \emph{arXiv preprint arXiv:2306.04140}.

\bibitem[{Dao et~al.(2023)Dao, Le, Phan, and Ngo}]{dao2023can}
Xuan-Quy Dao, Ngoc-Bich Le, Xuan-Dung Phan, and Bac-Bien Ngo. 2023.
\newblock Can chatgpt pass the vietnamese national high school graduation examination?
\newblock \emph{arXiv preprint arXiv:2306.09170}.

\bibitem[{Elkins et~al.(2023)Elkins, Kochmar, Serban, and Cheung}]{elkins2023useful}
Sabina Elkins, Ekaterina Kochmar, Iulian Serban, and Jackie~CK Cheung. 2023.
\newblock How useful are educational questions generated by large language models?
\newblock In \emph{International Conference on Artificial Intelligence in Education}, pages 536--542. Springer.

\bibitem[{Hendrycks et~al.(2020)Hendrycks, Burns, Basart, Zou, Mazeika, Song, and Steinhardt}]{hendrycks2020measuring}
Dan Hendrycks, Collin Burns, Steven Basart, Andy Zou, Mantas Mazeika, Dawn Song, and Jacob Steinhardt. 2020.
\newblock Measuring massive multitask language understanding.
\newblock \emph{arXiv preprint arXiv:2009.03300}.

\bibitem[{Huang et~al.(2023)Huang, Bai, Zhu, Zhang, Zhang, Su, Liu, Lv, Zhang, Lei et~al.}]{huang2023c}
Yuzhen Huang, Yuzhuo Bai, Zhihao Zhu, Junlei Zhang, Jinghan Zhang, Tangjun Su, Junteng Liu, Chuancheng Lv, Yikai Zhang, Jiayi Lei, et~al. 2023.
\newblock C-eval: A multi-level multi-discipline chinese evaluation suite for foundation models.
\newblock \emph{arXiv preprint arXiv:2305.08322}.

\bibitem[{Izacard and Grave(2020)}]{izacard2020leveraging}
Gautier Izacard and Edouard Grave. 2020.
\newblock Leveraging passage retrieval with generative models for open domain question answering.
\newblock \emph{arXiv preprint arXiv:2007.01282}.

\bibitem[{Kai et~al.(2021)Kai, Wu, Tang, Zhuang, Ding, Xiao, Long et~al.}]{kai2021learning}
Shen Kai, Lingfei Wu, Siliang Tang, Yueting Zhuang, Zhuoye Ding, Yun Xiao, Bo~Long, et~al. 2021.
\newblock Learning to generate visual questions with noisy supervision.
\newblock \emph{Advances in Neural Information Processing Systems}, 34:11604--11617.

\bibitem[{Kawabata and Sugawara(2023)}]{kawabata2023evaluating}
Akira Kawabata and Saku Sugawara. 2023.
\newblock Evaluating the rationale understanding of critical reasoning in logical reading comprehension.
\newblock In \emph{Proceedings of the 2023 Conference on Empirical Methods in Natural Language Processing}, pages 116--143.

\bibitem[{Kojima et~al.(2022)Kojima, Gu, Reid, Matsuo, and Iwasawa}]{kojima2022large}
Takeshi Kojima, Shixiang~Shane Gu, Machel Reid, Yutaka Matsuo, and Yusuke Iwasawa. 2022.
\newblock Large language models are zero-shot reasoners.
\newblock \emph{Advances in neural information processing systems}, 35:22199--22213.

\bibitem[{Li et~al.(2023)Li, Zhang, Koto, Yang, Zhao, Gong, Duan, and Baldwin}]{li2023cmmlu}
Haonan Li, Yixuan Zhang, Fajri Koto, Yifei Yang, Hai Zhao, Yeyun Gong, Nan Duan, and Timothy Baldwin. 2023.
\newblock Cmmlu: Measuring massive multitask language understanding in chinese.
\newblock \emph{arXiv preprint arXiv:2306.09212}.

\bibitem[{Liu et~al.(2023{\natexlab{a}})Liu, Jin, Ren, Yu, Dong, Peng, Zhang, Peng, Zhang, Lyu et~al.}]{liu2023m3ke}
Chuang Liu, Renren Jin, Yuqi Ren, Linhao Yu, Tianyu Dong, Xiaohan Peng, Shuting Zhang, Jianxiang Peng, Peiyi Zhang, Qingqing Lyu, et~al. 2023{\natexlab{a}}.
\newblock M3ke: A massive multi-level multi-subject knowledge evaluation benchmark for chinese large language models.
\newblock \emph{arXiv preprint arXiv:2305.10263}.

\bibitem[{Liu et~al.(2023{\natexlab{b}})Liu, Jin, Wang, Cheng, Dou, and Wen}]{liu2023reta}
Jiongnan Liu, Jiajie Jin, Zihan Wang, Jiehan Cheng, Zhicheng Dou, and Ji-Rong Wen. 2023{\natexlab{b}}.
\newblock Reta-llm: A retrieval-augmented large language model toolkit.
\newblock \emph{arXiv preprint arXiv:2306.05212}.

\bibitem[{OpenAI(2023)}]{openai2023GPT4}
OpenAI. 2023.
\newblock \href {http://arxiv.org/abs/2303.08774} {Gpt-4 technical report}.

\bibitem[{Raina and Gales(2022)}]{raina2022multiple}
Vatsal Raina and Mark Gales. 2022.
\newblock Multiple-choice question generation: Towards an automated assessment framework.
\newblock \emph{arXiv preprint arXiv:2209.11830}.

\bibitem[{Rajpurkar et~al.(2016)Rajpurkar, Zhang, Lopyrev, and Liang}]{rajpurkar2016squad}
Pranav Rajpurkar, Jian Zhang, Konstantin Lopyrev, and Percy Liang. 2016.
\newblock Squad: 100,000+ questions for machine comprehension of text.
\newblock \emph{arXiv preprint arXiv:1606.05250}.

\bibitem[{Saad-Falcon et~al.(2023)Saad-Falcon, Barrow, Siu, Nenkova, Rossi, and Dernoncourt}]{saad2023pdftriage}
Jon Saad-Falcon, Joe Barrow, Alexa Siu, Ani Nenkova, Ryan~A Rossi, and Franck Dernoncourt. 2023.
\newblock Pdftriage: Question answering over long, structured documents.
\newblock \emph{arXiv preprint arXiv:2309.08872}.

\bibitem[{Shao et~al.(2023)Shao, Li, Dai, and Qiu}]{shao2023character}
Yunfan Shao, Linyang Li, Junqi Dai, and Xipeng Qiu. 2023.
\newblock Character-llm: A trainable agent for role-playing.
\newblock \emph{arXiv preprint arXiv:2310.10158}.

\bibitem[{Tavares et~al.(2023)Tavares, Semedo, Rudnicky, and Magalhaes}]{tavares2023learning}
Diogo Tavares, David Semedo, Alexander Rudnicky, and Joao Magalhaes. 2023.
\newblock Learning to ask questions for zero-shot dialogue state tracking.
\newblock In \emph{Proceedings of the 46th International ACM SIGIR Conference on Research and Development in Information Retrieval}, pages 2118--2122.

\bibitem[{Touvron et~al.(2023)Touvron, Martin, Stone, Albert, Almahairi, Babaei, Bashlykov, Batra, Bhargava, Bhosale, Bikel, Blecher, Ferrer, Chen, Cucurull, Esiobu, Fernandes, Fu, Fu, Fuller, Gao, Goswami, Goyal, Hartshorn, Hosseini, Hou, Inan, Kardas, Kerkez, Khabsa, Kloumann, Korenev, Koura, Lachaux, Lavril, Lee, Liskovich, Lu, Mao, Martinet, Mihaylov, Mishra, Molybog, Nie, Poulton, Reizenstein, Rungta, Saladi, Schelten, Silva, Smith, Subramanian, Tan, Tang, Taylor, Williams, Kuan, Xu, Yan, Zarov, Zhang, Fan, Kambadur, Narang, Rodriguez, Stojnic, Edunov, and Scialom}]{touvron2023llama}
Hugo Touvron, Louis Martin, Kevin Stone, Peter Albert, Amjad Almahairi, Yasmine Babaei, Nikolay Bashlykov, Soumya Batra, Prajjwal Bhargava, Shruti Bhosale, Dan Bikel, Lukas Blecher, Cristian~Canton Ferrer, Moya Chen, Guillem Cucurull, David Esiobu, Jude Fernandes, Jeremy Fu, Wenyin Fu, Brian Fuller, Cynthia Gao, Vedanuj Goswami, Naman Goyal, Anthony Hartshorn, Saghar Hosseini, Rui Hou, Hakan Inan, Marcin Kardas, Viktor Kerkez, Madian Khabsa, Isabel Kloumann, Artem Korenev, Punit~Singh Koura, Marie-Anne Lachaux, Thibaut Lavril, Jenya Lee, Diana Liskovich, Yinghai Lu, Yuning Mao, Xavier Martinet, Todor Mihaylov, Pushkar Mishra, Igor Molybog, Yixin Nie, Andrew Poulton, Jeremy Reizenstein, Rashi Rungta, Kalyan Saladi, Alan Schelten, Ruan Silva, Eric~Michael Smith, Ranjan Subramanian, Xiaoqing~Ellen Tan, Binh Tang, Ross Taylor, Adina Williams, Jian~Xiang Kuan, Puxin Xu, Zheng Yan, Iliyan Zarov, Yuchen Zhang, Angela Fan, Melanie Kambadur, Sharan Narang, Aurelien Rodriguez, Robert Stojnic, Sergey Edunov, and Thomas
  Scialom. 2023.
\newblock \href {http://arxiv.org/abs/2307.09288} {Llama 2: Open foundation and fine-tuned chat models}.

\bibitem[{Tran and Kretchmar(2023)}]{tran2023single}
Son~Quoc Tran and Matt Kretchmar. 2023.
\newblock Single-sentence reader: A novel approach for addressing answer position bias.
\newblock \emph{arXiv preprint arXiv:2308.04566}.

\bibitem[{Uehara et~al.(2022)Uehara, Duan, and Harada}]{uehara2022learning}
Kohei Uehara, Nan Duan, and Tatsuya Harada. 2022.
\newblock Learning to ask informative sub-questions for visual question answering.
\newblock In \emph{Proceedings of the IEEE/CVF Conference on Computer Vision and Pattern Recognition}, pages 4681--4690.

\bibitem[{Wang et~al.(2023)Wang, Hu, Lu, Zhu, Zhang, Subramaniam, Loomba, Zhang, Sun, and Wang}]{wang2023scibench}
Xiaoxuan Wang, Ziniu Hu, Pan Lu, Yanqiao Zhu, Jieyu Zhang, Satyen Subramaniam, Arjun~R Loomba, Shichang Zhang, Yizhou Sun, and Wei Wang. 2023.
\newblock Scibench: Evaluating college-level scientific problem-solving abilities of large language models.
\newblock \emph{arXiv preprint arXiv:2307.10635}.

\bibitem[{Wei et~al.(2023)Wei, Luan, Liu, Dong, and Wang}]{wei2023cmath}
Tianwen Wei, Jian Luan, Wei Liu, Shuang Dong, and Bin Wang. 2023.
\newblock Cmath: Can your language model pass chinese elementary school math test?
\newblock \emph{arXiv preprint arXiv:2306.16636}.

\bibitem[{Workshop et~al.(2023)Workshop, :, Scao, Fan, Akiki, Pavlick, Ilić, Hesslow, Castagné, Luccioni, Yvon, Gallé, Tow, Rush, Biderman, Webson, Ammanamanchi, Wang, Sagot, Muennighoff, del Moral, Ruwase, Bawden, Bekman, McMillan-Major, Beltagy, Nguyen, Saulnier, Tan, Suarez, Sanh, Laurençon, Jernite, Launay, Mitchell, Raffel, Gokaslan, Simhi, Soroa, Aji, Alfassy, Rogers, Nitzav, Xu, Mou, Emezue, Klamm, Leong, van Strien, Adelani, Radev, Ponferrada, Levkovizh, Kim, Natan, Toni, Dupont, Kruszewski, Pistilli, Elsahar, Benyamina, Tran, Yu, Abdulmumin, Johnson, Gonzalez-Dios, de~la Rosa, Chim, Dodge, Zhu, Chang, Frohberg, Tobing, Bhattacharjee, Almubarak, Chen, Lo, Werra, Weber, Phan, allal, Tanguy, Dey, Muñoz, Masoud, Grandury, Šaško, Huang, Coavoux, Singh, Jiang, Vu, Jauhar, Ghaleb, Subramani, Kassner, Khamis, Nguyen, Espejel, de~Gibert, Villegas, Henderson, Colombo, Amuok, Lhoest, Harliman, Bommasani, López, Ribeiro, Osei, Pyysalo, Nagel, Bose, Muhammad, Sharma, Longpre, Nikpoor, Silberberg, Pai,
  Zink, Torrent, Schick, Thrush, Danchev, Nikoulina, Laippala, Lepercq, Prabhu, Alyafeai, Talat, Raja, Heinzerling, Si, Taşar, Salesky, Mielke, Lee, Sharma, Santilli, Chaffin, Stiegler, Datta, Szczechla, Chhablani, Wang, Pandey, Strobelt, Fries, Rozen, Gao, Sutawika, Bari, Al-shaibani, Manica, Nayak, Teehan, Albanie, Shen, Ben-David, Bach, Kim, Bers, Fevry, Neeraj, Thakker, Raunak, Tang, Yong, Sun, Brody, Uri, Tojarieh, Roberts, Chung, Tae, Phang, Press, Li, Narayanan, Bourfoune, Casper, Rasley, Ryabinin, Mishra, Zhang, Shoeybi, Peyrounette, Patry, Tazi, Sanseviero, von Platen, Cornette, Lavallée, Lacroix, Rajbhandari, Gandhi, Smith, Requena, Patil, Dettmers, Baruwa, Singh, Cheveleva, Ligozat, Subramonian, Névéol, Lovering, Garrette, Tunuguntla, Reiter, Taktasheva, Voloshina, Bogdanov, Winata, Schoelkopf, Kalo, Novikova, Forde, Clive, Kasai, Kawamura, Hazan, Carpuat, Clinciu, Kim, Cheng, Serikov, Antverg, van~der Wal, Zhang, Zhang, Gehrmann, Mirkin, Pais, Shavrina, Scialom, Yun, Limisiewicz, Rieser,
  Protasov, Mikhailov, Pruksachatkun, Belinkov, Bamberger, Kasner, Rueda, Pestana, Feizpour, Khan, Faranak, Santos, Hevia, Unldreaj, Aghagol, Abdollahi, Tammour, HajiHosseini, Behroozi, Ajibade, Saxena, Ferrandis, McDuff, Contractor, Lansky, David, Kiela, Nguyen, Tan, Baylor, Ozoani, Mirza, Ononiwu, Rezanejad, Jones, Bhattacharya, Solaiman, Sedenko, Nejadgholi, Passmore, Seltzer, Sanz, Dutra, Samagaio, Elbadri, Mieskes, Gerchick, Akinlolu, McKenna, Qiu, Ghauri, Burynok, Abrar, Rajani, Elkott, Fahmy, Samuel, An, Kromann, Hao, Alizadeh, Shubber, Wang, Roy, Viguier, Le, Oyebade, Le, Yang, Nguyen, Kashyap, Palasciano, Callahan, Shukla, Miranda-Escalada, Singh, Beilharz, Wang, Brito, Zhou, Jain, Xu, Fourrier, Periñán, Molano, Yu, Manjavacas, Barth, Fuhrimann, Altay, Bayrak, Burns, Vrabec, Bello, Dash, Kang, Giorgi, Golde, Posada, Sivaraman, Bulchandani, Liu, Shinzato, de~Bykhovetz, Takeuchi, Pàmies, Castillo, Nezhurina, Sänger, Samwald, Cullan, Weinberg, Wolf, Mihaljcic, Liu, Freidank, Kang, Seelam, Dahlberg,
  Broad, Muellner, Fung, Haller, Chandrasekhar, Eisenberg, Martin, Canalli, Su, Su, Cahyawijaya, Garda, Deshmukh, Mishra, Kiblawi, Ott, Sang-aroonsiri, Kumar, Schweter, Bharati, Laud, Gigant, Kainuma, Kusa, Labrak, Bajaj, Venkatraman, Xu, Xu, Xu, Tan, Xie, Ye, Bras, Belkada, and Wolf}]{workshop2023bloom}
BigScience Workshop, :, Teven~Le Scao, Angela Fan, Christopher Akiki, Ellie Pavlick, Suzana Ilić, Daniel Hesslow, Roman Castagné, Alexandra~Sasha Luccioni, François Yvon, Matthias Gallé, Jonathan Tow, Alexander~M. Rush, Stella Biderman, Albert Webson, Pawan~Sasanka Ammanamanchi, Thomas Wang, Benoît Sagot, Niklas Muennighoff, Albert~Villanova del Moral, Olatunji Ruwase, Rachel Bawden, Stas Bekman, Angelina McMillan-Major, Iz~Beltagy, Huu Nguyen, Lucile Saulnier, Samson Tan, Pedro~Ortiz Suarez, Victor Sanh, Hugo Laurençon, Yacine Jernite, Julien Launay, Margaret Mitchell, Colin Raffel, Aaron Gokaslan, Adi Simhi, Aitor Soroa, Alham~Fikri Aji, Amit Alfassy, Anna Rogers, Ariel~Kreisberg Nitzav, Canwen Xu, Chenghao Mou, Chris Emezue, Christopher Klamm, Colin Leong, Daniel van Strien, David~Ifeoluwa Adelani, Dragomir Radev, Eduardo~González Ponferrada, Efrat Levkovizh, Ethan Kim, Eyal~Bar Natan, Francesco~De Toni, Gérard Dupont, Germán Kruszewski, Giada Pistilli, Hady Elsahar, Hamza Benyamina, Hieu Tran,
  Ian Yu, Idris Abdulmumin, Isaac Johnson, Itziar Gonzalez-Dios, Javier de~la Rosa, Jenny Chim, Jesse Dodge, Jian Zhu, Jonathan Chang, Jörg Frohberg, Joseph Tobing, Joydeep Bhattacharjee, Khalid Almubarak, Kimbo Chen, Kyle Lo, Leandro~Von Werra, Leon Weber, Long Phan, Loubna~Ben allal, Ludovic Tanguy, Manan Dey, Manuel~Romero Muñoz, Maraim Masoud, María Grandury, Mario Šaško, Max Huang, Maximin Coavoux, Mayank Singh, Mike Tian-Jian Jiang, Minh~Chien Vu, Mohammad~A. Jauhar, Mustafa Ghaleb, Nishant Subramani, Nora Kassner, Nurulaqilla Khamis, Olivier Nguyen, Omar Espejel, Ona de~Gibert, Paulo Villegas, Peter Henderson, Pierre Colombo, Priscilla Amuok, Quentin Lhoest, Rheza Harliman, Rishi Bommasani, Roberto~Luis López, Rui Ribeiro, Salomey Osei, Sampo Pyysalo, Sebastian Nagel, Shamik Bose, Shamsuddeen~Hassan Muhammad, Shanya Sharma, Shayne Longpre, Somaieh Nikpoor, Stanislav Silberberg, Suhas Pai, Sydney Zink, Tiago~Timponi Torrent, Timo Schick, Tristan Thrush, Valentin Danchev, Vassilina Nikoulina,
  Veronika Laippala, Violette Lepercq, Vrinda Prabhu, Zaid Alyafeai, Zeerak Talat, Arun Raja, Benjamin Heinzerling, Chenglei Si, Davut~Emre Taşar, Elizabeth Salesky, Sabrina~J. Mielke, Wilson~Y. Lee, Abheesht Sharma, Andrea Santilli, Antoine Chaffin, Arnaud Stiegler, Debajyoti Datta, Eliza Szczechla, Gunjan Chhablani, Han Wang, Harshit Pandey, Hendrik Strobelt, Jason~Alan Fries, Jos Rozen, Leo Gao, Lintang Sutawika, M~Saiful Bari, Maged~S. Al-shaibani, Matteo Manica, Nihal Nayak, Ryan Teehan, Samuel Albanie, Sheng Shen, Srulik Ben-David, Stephen~H. Bach, Taewoon Kim, Tali Bers, Thibault Fevry, Trishala Neeraj, Urmish Thakker, Vikas Raunak, Xiangru Tang, Zheng-Xin Yong, Zhiqing Sun, Shaked Brody, Yallow Uri, Hadar Tojarieh, Adam Roberts, Hyung~Won Chung, Jaesung Tae, Jason Phang, Ofir Press, Conglong Li, Deepak Narayanan, Hatim Bourfoune, Jared Casper, Jeff Rasley, Max Ryabinin, Mayank Mishra, Minjia Zhang, Mohammad Shoeybi, Myriam Peyrounette, Nicolas Patry, Nouamane Tazi, Omar Sanseviero, Patrick von
  Platen, Pierre Cornette, Pierre~François Lavallée, Rémi Lacroix, Samyam Rajbhandari, Sanchit Gandhi, Shaden Smith, Stéphane Requena, Suraj Patil, Tim Dettmers, Ahmed Baruwa, Amanpreet Singh, Anastasia Cheveleva, Anne-Laure Ligozat, Arjun Subramonian, Aurélie Névéol, Charles Lovering, Dan Garrette, Deepak Tunuguntla, Ehud Reiter, Ekaterina Taktasheva, Ekaterina Voloshina, Eli Bogdanov, Genta~Indra Winata, Hailey Schoelkopf, Jan-Christoph Kalo, Jekaterina Novikova, Jessica~Zosa Forde, Jordan Clive, Jungo Kasai, Ken Kawamura, Liam Hazan, Marine Carpuat, Miruna Clinciu, Najoung Kim, Newton Cheng, Oleg Serikov, Omer Antverg, Oskar van~der Wal, Rui Zhang, Ruochen Zhang, Sebastian Gehrmann, Shachar Mirkin, Shani Pais, Tatiana Shavrina, Thomas Scialom, Tian Yun, Tomasz Limisiewicz, Verena Rieser, Vitaly Protasov, Vladislav Mikhailov, Yada Pruksachatkun, Yonatan Belinkov, Zachary Bamberger, Zdeněk Kasner, Alice Rueda, Amanda Pestana, Amir Feizpour, Ammar Khan, Amy Faranak, Ana Santos, Anthony Hevia, Antigona
  Unldreaj, Arash Aghagol, Arezoo Abdollahi, Aycha Tammour, Azadeh HajiHosseini, Bahareh Behroozi, Benjamin Ajibade, Bharat Saxena, Carlos~Muñoz Ferrandis, Daniel McDuff, Danish Contractor, David Lansky, Davis David, Douwe Kiela, Duong~A. Nguyen, Edward Tan, Emi Baylor, Ezinwanne Ozoani, Fatima Mirza, Frankline Ononiwu, Habib Rezanejad, Hessie Jones, Indrani Bhattacharya, Irene Solaiman, Irina Sedenko, Isar Nejadgholi, Jesse Passmore, Josh Seltzer, Julio~Bonis Sanz, Livia Dutra, Mairon Samagaio, Maraim Elbadri, Margot Mieskes, Marissa Gerchick, Martha Akinlolu, Michael McKenna, Mike Qiu, Muhammed Ghauri, Mykola Burynok, Nafis Abrar, Nazneen Rajani, Nour Elkott, Nour Fahmy, Olanrewaju Samuel, Ran An, Rasmus Kromann, Ryan Hao, Samira Alizadeh, Sarmad Shubber, Silas Wang, Sourav Roy, Sylvain Viguier, Thanh Le, Tobi Oyebade, Trieu Le, Yoyo Yang, Zach Nguyen, Abhinav~Ramesh Kashyap, Alfredo Palasciano, Alison Callahan, Anima Shukla, Antonio Miranda-Escalada, Ayush Singh, Benjamin Beilharz, Bo~Wang, Caio Brito,
  Chenxi Zhou, Chirag Jain, Chuxin Xu, Clémentine Fourrier, Daniel~León Periñán, Daniel Molano, Dian Yu, Enrique Manjavacas, Fabio Barth, Florian Fuhrimann, Gabriel Altay, Giyaseddin Bayrak, Gully Burns, Helena~U. Vrabec, Imane Bello, Ishani Dash, Jihyun Kang, John Giorgi, Jonas Golde, Jose~David Posada, Karthik~Rangasai Sivaraman, Lokesh Bulchandani, Lu~Liu, Luisa Shinzato, Madeleine~Hahn de~Bykhovetz, Maiko Takeuchi, Marc Pàmies, Maria~A Castillo, Marianna Nezhurina, Mario Sänger, Matthias Samwald, Michael Cullan, Michael Weinberg, Michiel~De Wolf, Mina Mihaljcic, Minna Liu, Moritz Freidank, Myungsun Kang, Natasha Seelam, Nathan Dahlberg, Nicholas~Michio Broad, Nikolaus Muellner, Pascale Fung, Patrick Haller, Ramya Chandrasekhar, Renata Eisenberg, Robert Martin, Rodrigo Canalli, Rosaline Su, Ruisi Su, Samuel Cahyawijaya, Samuele Garda, Shlok~S Deshmukh, Shubhanshu Mishra, Sid Kiblawi, Simon Ott, Sinee Sang-aroonsiri, Srishti Kumar, Stefan Schweter, Sushil Bharati, Tanmay Laud, Théo Gigant, Tomoya
  Kainuma, Wojciech Kusa, Yanis Labrak, Yash~Shailesh Bajaj, Yash Venkatraman, Yifan Xu, Yingxin Xu, Yu~Xu, Zhe Tan, Zhongli Xie, Zifan Ye, Mathilde Bras, Younes Belkada, and Thomas Wolf. 2023.
\newblock \href {http://arxiv.org/abs/2211.05100} {Bloom: A 176b-parameter open-access multilingual language model}.

\bibitem[{Zeng(2023)}]{zeng2023measuring}
Hui Zeng. 2023.
\newblock Measuring massive multitask chinese understanding.
\newblock \emph{arXiv preprint arXiv:2304.12986}.

\bibitem[{Zhang et~al.(2023{\natexlab{a}})Zhang, Cai, Liu, Yang, Dai, Liao, Qin, Li, Liu, Liu et~al.}]{zhang2023fineval}
Liwen Zhang, Weige Cai, Zhaowei Liu, Zhi Yang, Wei Dai, Yujie Liao, Qianru Qin, Yifei Li, Xingyu Liu, Zhiqiang Liu, et~al. 2023{\natexlab{a}}.
\newblock Fineval: A chinese financial domain knowledge evaluation benchmark for large language models.
\newblock \emph{arXiv preprint arXiv:2308.09975}.

\bibitem[{Zhang et~al.(2023{\natexlab{b}})Zhang, Li, Zong, Ying, He, and Qiu}]{zhang2023evaluating}
Xiaotian Zhang, Chunyang Li, Yi~Zong, Zhengyu Ying, Liang He, and Xipeng Qiu. 2023{\natexlab{b}}.
\newblock Evaluating the performance of large language models on gaokao benchmark.
\newblock \emph{arXiv preprint arXiv:2305.12474}.

\bibitem[{Zheng et~al.(2023)Zheng, Chiang, Sheng, Zhuang, Wu, Zhuang, Lin, Li, Li, Xing, Zhang, Gonzalez, and Stoica}]{zheng2023judging}
Lianmin Zheng, Wei-Lin Chiang, Ying Sheng, Siyuan Zhuang, Zhanghao Wu, Yonghao Zhuang, Zi~Lin, Zhuohan Li, Dacheng Li, Eric.~P Xing, Hao Zhang, Joseph~E. Gonzalez, and Ion Stoica. 2023.
\newblock \href {http://arxiv.org/abs/2306.05685} {Judging llm-as-a-judge with mt-bench and chatbot arena}.

\bibitem[{Zhong et~al.(2023)Zhong, Cui, Guo, Liang, Lu, Wang, Saied, Chen, and Duan}]{zhong2023agieval}
Wanjun Zhong, Ruixiang Cui, Yiduo Guo, Yaobo Liang, Shuai Lu, Yanlin Wang, Amin Saied, Weizhu Chen, and Nan Duan. 2023.
\newblock Agieval: A human-centric benchmark for evaluating foundation models.
\newblock \emph{arXiv preprint arXiv:2304.06364}.

\bibitem[{Zhou et~al.(2023{\natexlab{a}})Zhou, Zhang, Cao, Li, and Wang}]{zhou2023complementary}
Tongquan Zhou, Yao Zhang, Siyi Cao, Yulu Li, and Tao Wang. 2023{\natexlab{a}}.
\newblock Complementary advantages of chatgpts and human readers in reasoning: Evidence from english text reading comprehension.
\newblock \emph{arXiv preprint arXiv:2311.10344}.

\bibitem[{Zhou et~al.(2023{\natexlab{b}})Zhou, Geng, Shen, Tao, Long, Lou, and Shen}]{zhou2023thread}
Yucheng Zhou, Xiubo Geng, Tao Shen, Chongyang Tao, Guodong Long, Jian-Guang Lou, and Jianbing Shen. 2023{\natexlab{b}}.
\newblock Thread of thought unraveling chaotic contexts.
\newblock \emph{arXiv preprint arXiv:2311.08734}.

\end{thebibliography}

\appendix

\begin{table*}[]
\centering
\resizebox{0.9\textwidth}{!}{%
\begin{tabular}{p{4cm}|p{2cm}|p{14cm}}
\toprule
\multirow{2}{*}{General/Monodisciplinary} & Generation & Given a text, please propose some questions for students. These questions are divided into six levels: memory, understanding, application, analysis, evaluation, and creation.The memory level requires students to recall specific facts or information from the article. For example, 'What are the main ecosystem services mentioned in the article?';The understanding level checks whether students have grasped the concepts or arguments in the article. For example, 'Explain what biodiversity is and why it is important to ecosystems.';The application level involves applying understood concepts to new situations or examples. For example, 'How would the reduction in biodiversity affect ecosystem services?';The analysis level involves dissecting the information in the article, such as comparing and contrasting different concepts or situations. For example, 'Compare the importance of different ecosystem services mentioned in the article.';The evaluation level assesses or critiques the views, methods, or arguments of the article. For example, 'Evaluate the author's argument about the importance of biodiversity conservation';The creation level involves combining different parts of the article to form new insights or summaries. For example, 'Integrate the information mentioned in the article to discuss the best strategy for biodiversity conservation.'Text content is xxx. You need to propose three questions of each level based on the above text. \\ \cmidrule(l){2-3} 
                                          & Evaluation & The text content is xxx. A question is xxx. You need to evaluate this question considering the following four criteria: i) Consistency. 0: The question does not align with any of the six educational levels (memory, understanding, application, analysis, evaluation, creation) as outlined in Anderson and Krathwohl's revised version of Bloom's Taxonomy.1: The question aligns with one of the six educational levels (memory, understanding, application, analysis, evaluation, creation) as mentioned in Anderson and Krathwohl's revised version of Bloom's Taxonomy. ii) Relevance. 0: The question is less than 50\% relevant to the provided textual context.1: The question is more than 50\% relevant to the provided textual context.iii) Coverage. 0: The content addressed by all questions does not cover more than 50\% of the textual context.1: The content addressed by all questions covers more than 50\% of the textual context. iv) Representativeness. 0: The question does not represent more than 50\% of the important content within the context. 1: The question represents more than 50\% of the important content within the context.For each criterion, you only need to give a number, where 1 represents 'meets the criterion' and 0 represents 'does not meet the criterion'.Give the result in this format: {[}x, x, x, x{]}                                                                                                                                                                    \\ \midrule
\multirow{2}{*}{Interdisciplinary}        & Generation & Create an interdisciplinary problem of subjectA and subjectB. Give detailed subjectA's knowledge as background and infer subjectB's events or phenomenon from subjectA's background. For example, "The Strait of Gibraltar is a narrow waterway that serves as a natural separation between the Iberian Peninsula in Europe and the North African coast. This strategic location has historically been a crucial passage for naval trade and military movements, influencing the development and outcome of various historical events in the Mediterranean region. Based on the geographical background provided about the Strait of Gibraltar, which of the following historical events or phenomena can be inferred to have been significantly influenced by its location?A) The establishment of the Silk RoadB) The expansion of the Roman Empire into Northern AfricaC) The Viking exploration of the North AtlanticD) The trade routes during the Age of Discovery"The subjectA and subjectB can be geography and history, physics and chemistry, biology and chemistry, art and history, physics and history, etc.Please give such a multiple-choice question.                                                                                                                                                                                                                                                                                                                                       \\\cmidrule(l){2-3}
                                          & Evaluation & A question is xxx. You need to evaluate this question considering the following two criteria: i) Relevance. 0: The question is less than 50\% relevant to the provided textual context. 1: The question is more than 50\% relevant to the provided textual context. ii) Representativeness. 0: The question does not represent more than 50\% of the important content within the context. 1: The question represents more than 50\% of the important content within the context. For each criterion, you only need to give a number, where 1 represents 'meets the criterion' and 0 represents 'does not meet the criterion'.Give the result in this format: {[}x, x{]}                                                                                                                                                                                                                                                                                                                                                                                                                                                                                                                                                                                                                                                                                                                                                                                                                                                    \\ \bottomrule 
\end{tabular}%
}
\caption{The instruction prompts of three domains in guiding LLMs to generate questions and making evaluation.}
\label{tab:xuexi-prompt}
\end{table*}

\begin{figure*}[!h]
  \centering
  \includegraphics[width=0.8\linewidth]{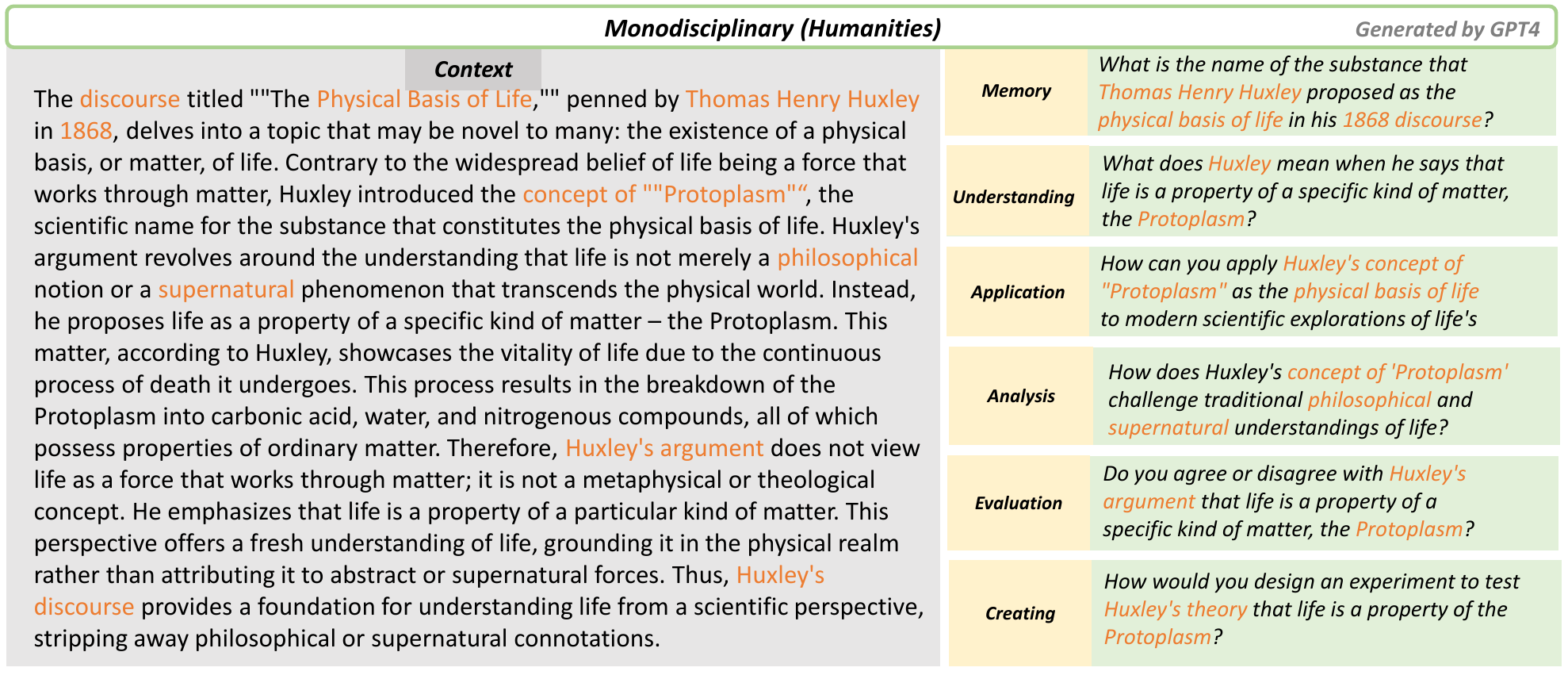}
  \caption{Questions posed by the top LLM, i.e. GPT4, in the humanities-related monodisciplinary domain. }
  \label{fig:xuexi-case2}
\end{figure*}

\begin{figure*}[!h]
  \centering
  \includegraphics[width=0.8\linewidth]{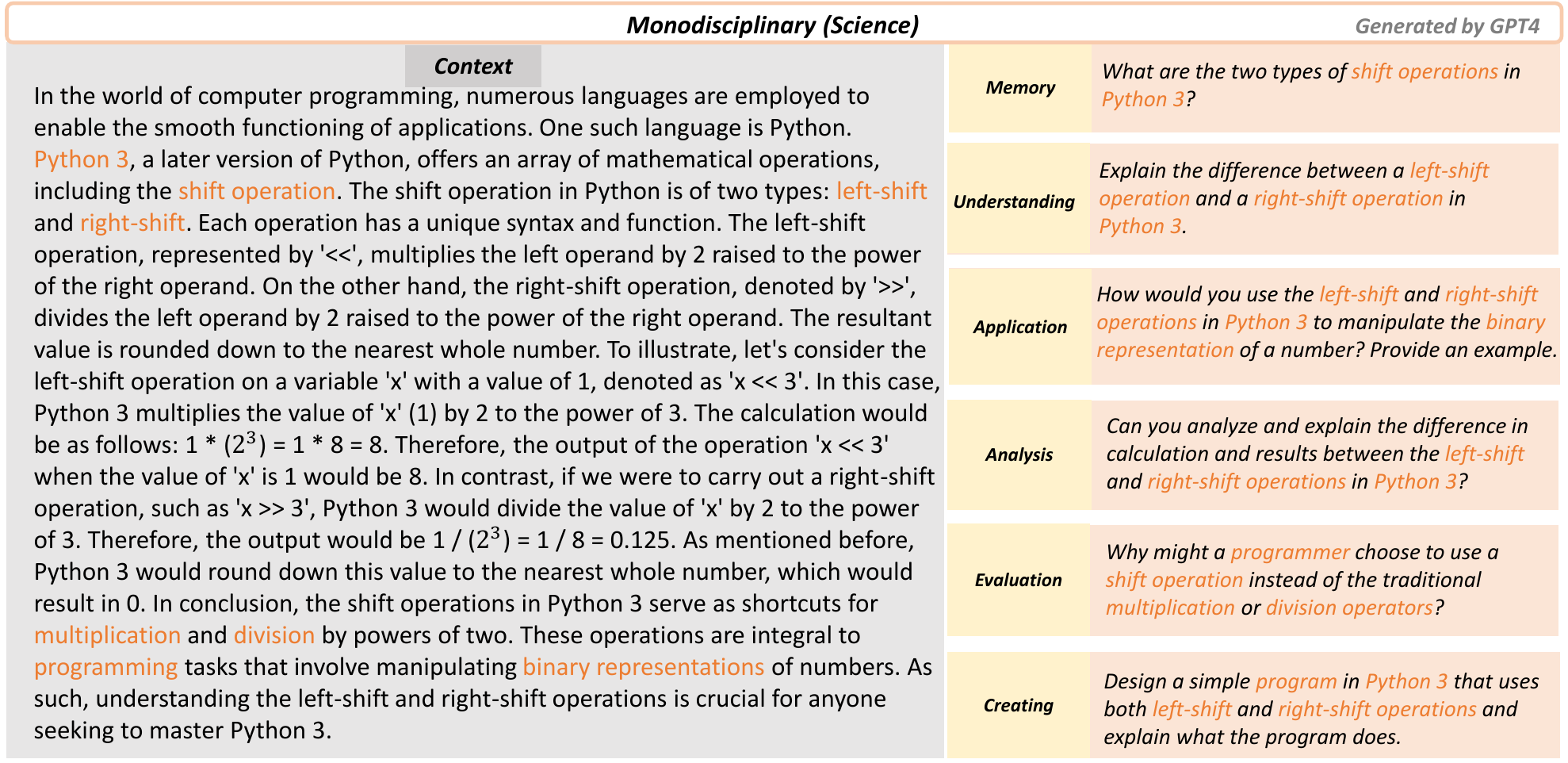}
  \caption{Questions posed by the top LLM, i.e. GPT4, in the science-related monodisciplinary domain.}
  \label{fig:xuexi-case3}
\end{figure*}

\begin{figure*}[!h]
  \centering
  \includegraphics[width=0.8\linewidth]{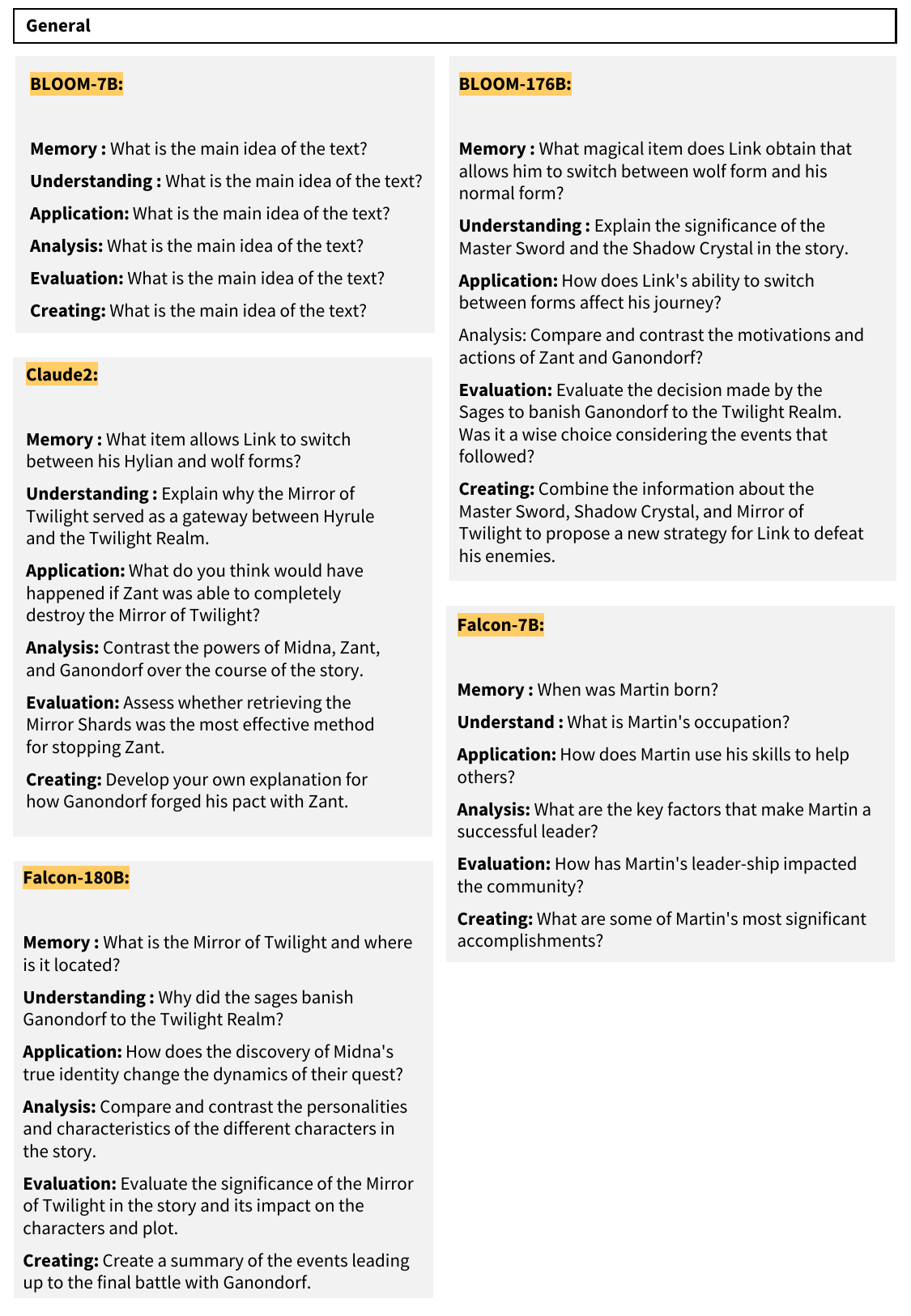}
  \caption{Questions posed by different LLMs in the general domain.}
  \label{fig:1xuexi-case1}
\end{figure*}

\begin{figure*}[!h]
  \centering
  \includegraphics[width=0.8\linewidth]{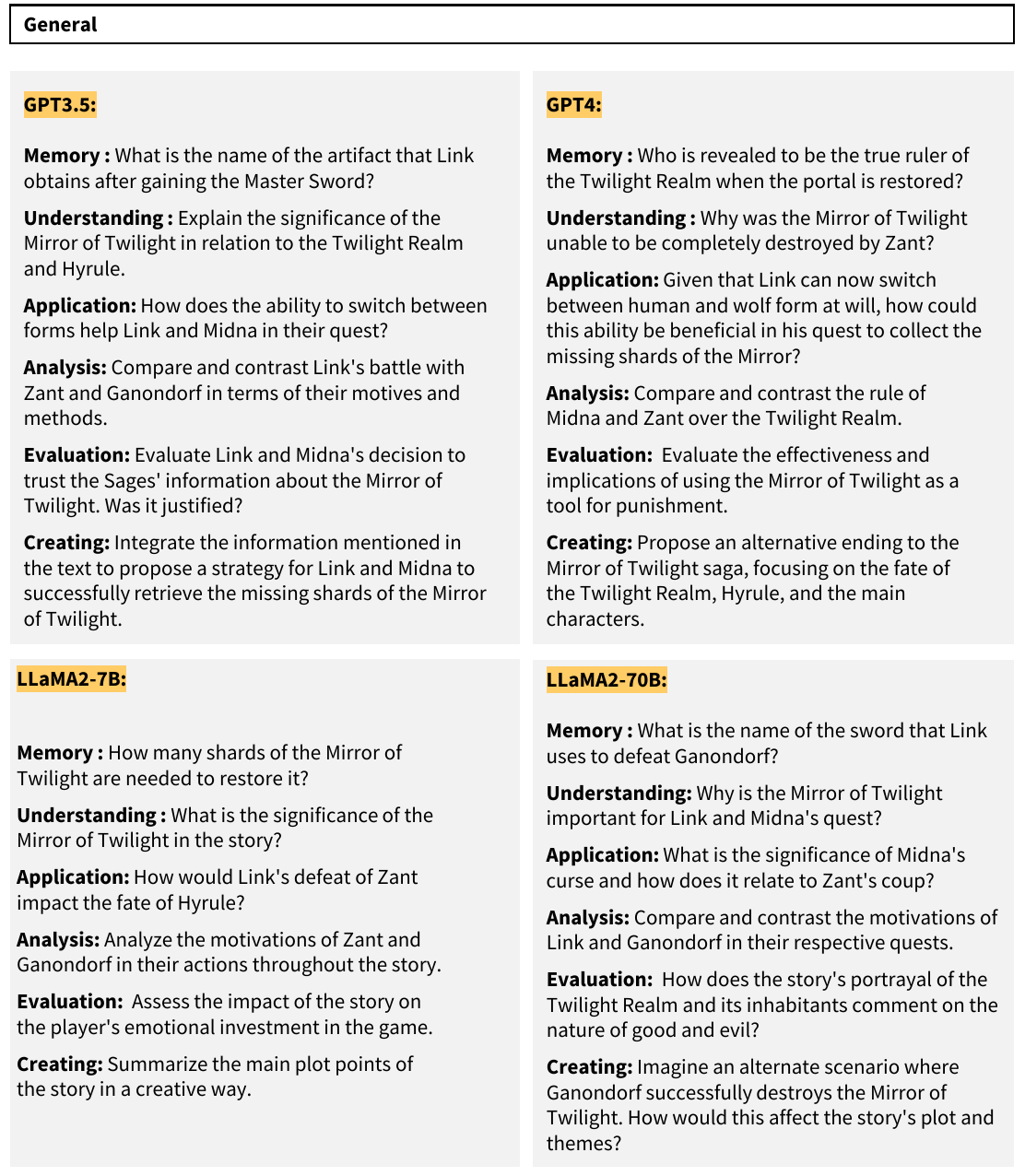}
  \caption{Questions posed by different LLMs in the general domain.}
  \label{fig:1xuexi-case2}
\end{figure*}

\begin{figure*}[!h]
  \centering
  \includegraphics[width=0.8\linewidth]{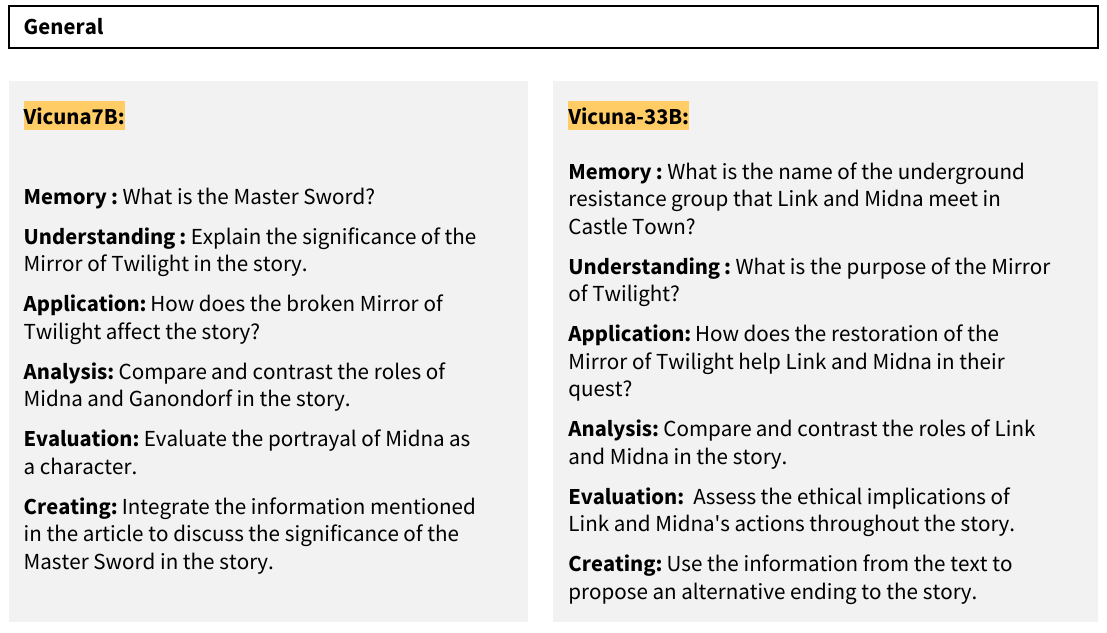}
  \caption{Questions posed by different LLMs in the general domain.}
  \label{fig:1xuexi-case3}
\end{figure*}

\begin{figure*}[!h]
  \centering
  \includegraphics[width=0.8\linewidth]{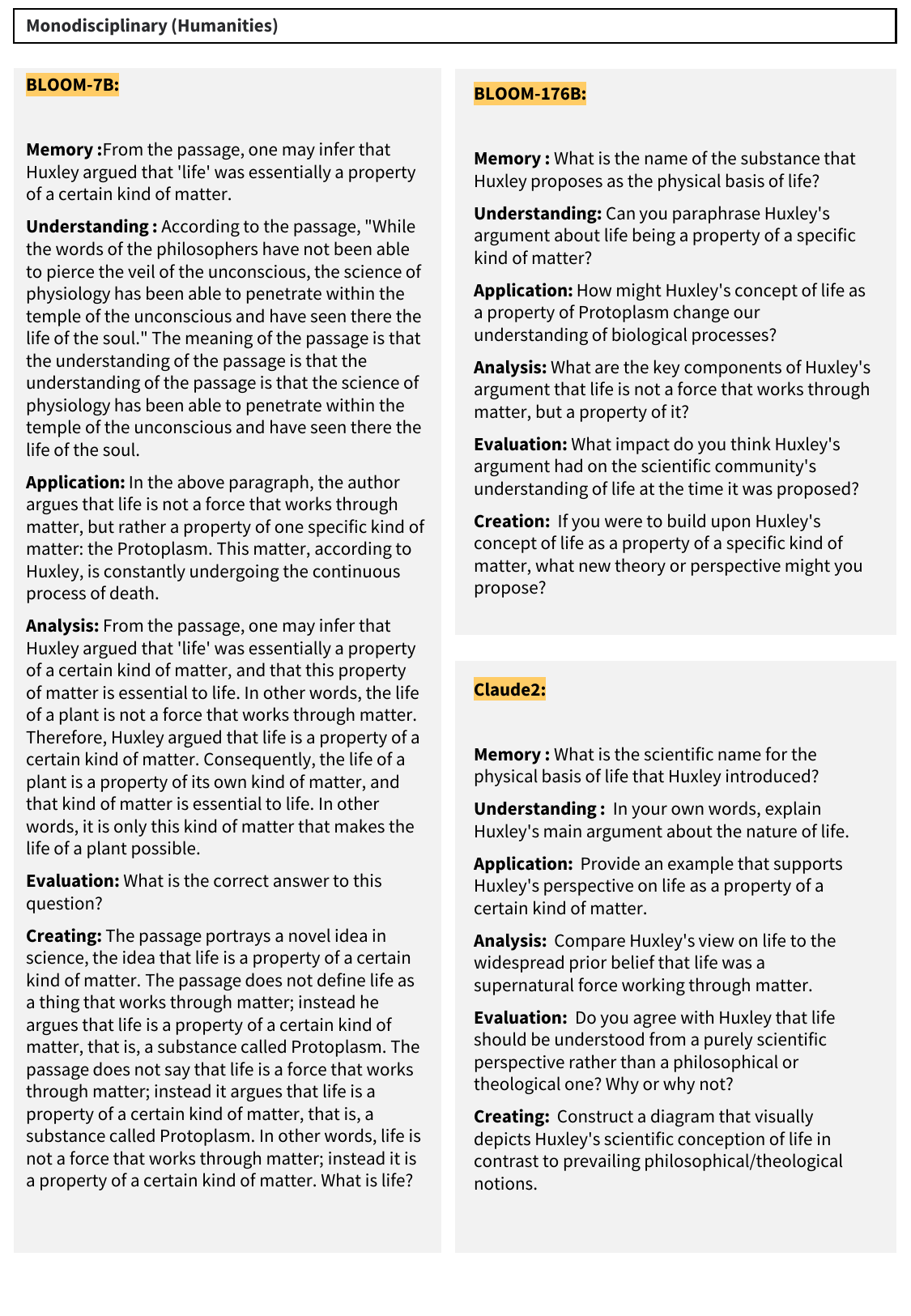}
  \caption{Questions posed by different LLMs in the humanities-related monodisciplinary domain. }
  \label{fig:2xuexi-case1}
\end{figure*}

\begin{figure*}[!h]
  \centering
  \includegraphics[width=0.8\linewidth]{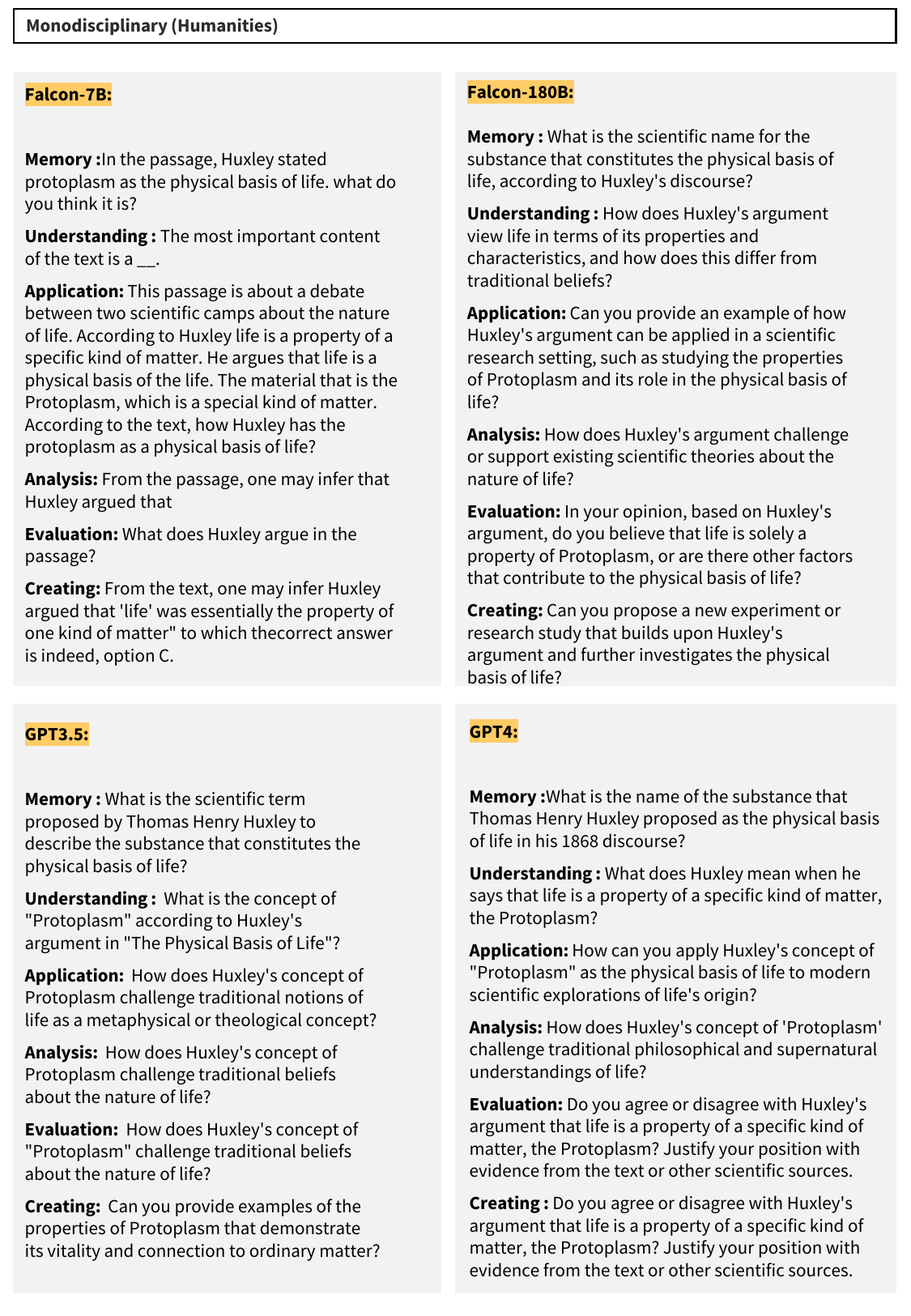}
  \caption{Questions posed by different LLMs in the humanities-related monodisciplinary domain. }
  \label{fig:2xuexi-case2}
\end{figure*}

\begin{figure*}[!h]
  \centering
  \includegraphics[width=0.8\linewidth]{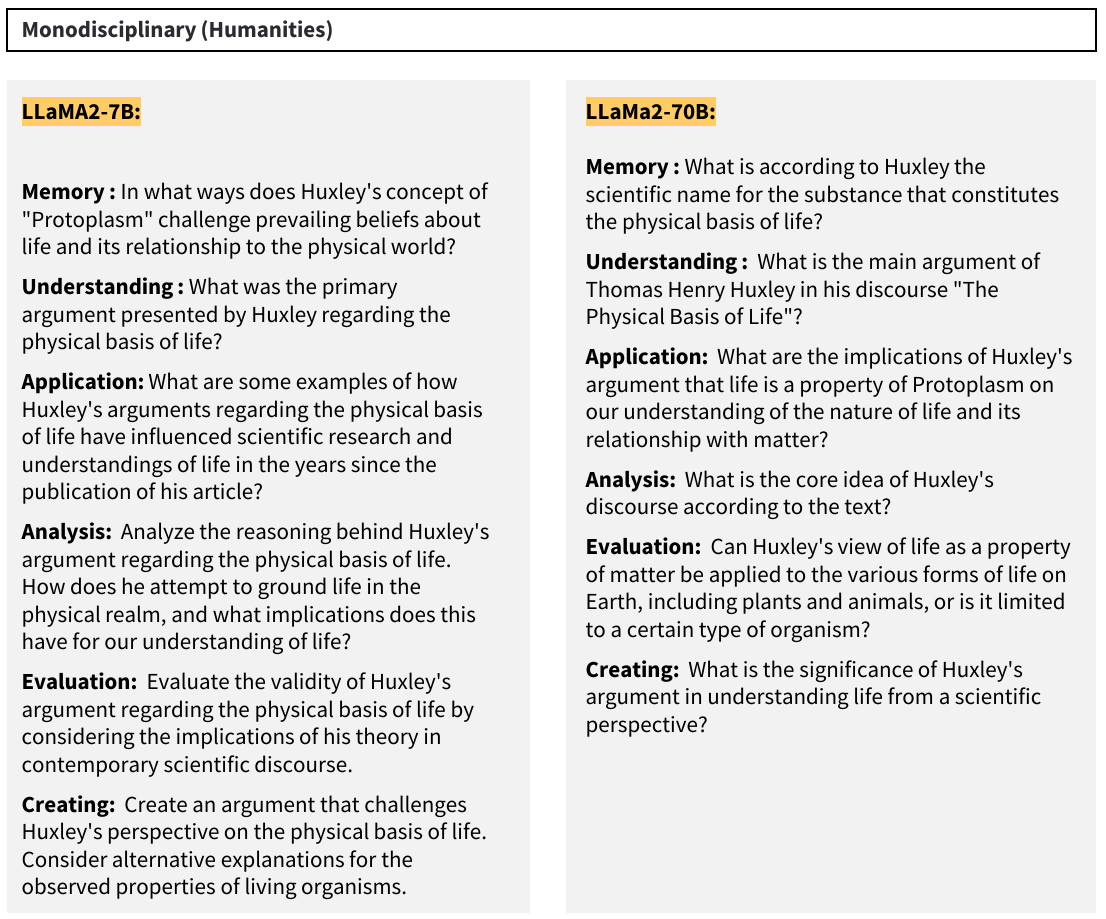}
  \caption{Questions posed by different LLMs in the humanities-related monodisciplinary domain. }
  \label{fig:2xuexi-case3}
\end{figure*}

\begin{figure*}[!h]
  \centering
  \includegraphics[width=0.8\linewidth]{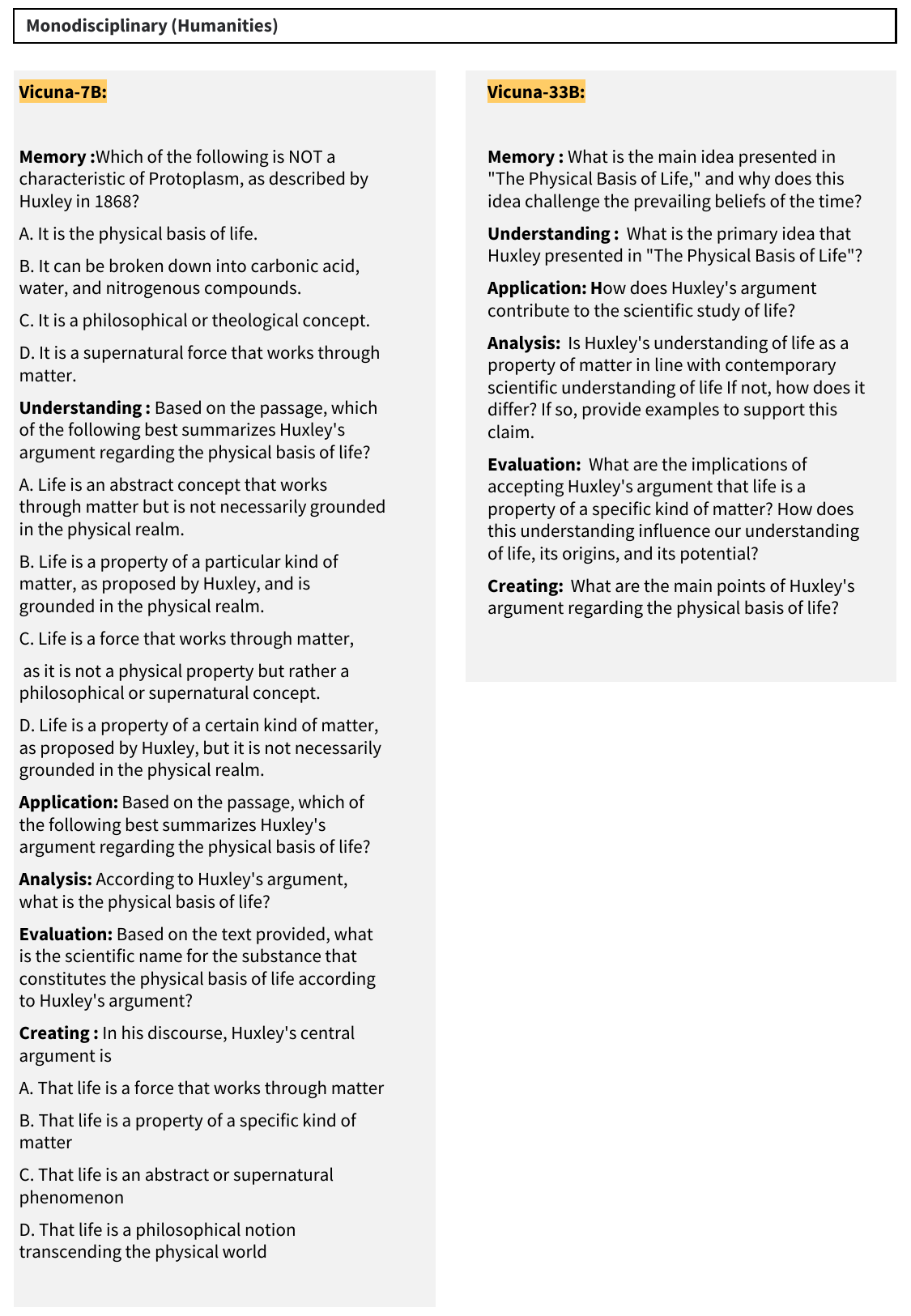}
  \caption{Questions posed by different LLMs in the humanities-related monodisciplinary domain. }
  \label{fig:2xuexi-case4}
\end{figure*}

\begin{figure*}[!h]
  \centering
  \includegraphics[width=0.8\linewidth]{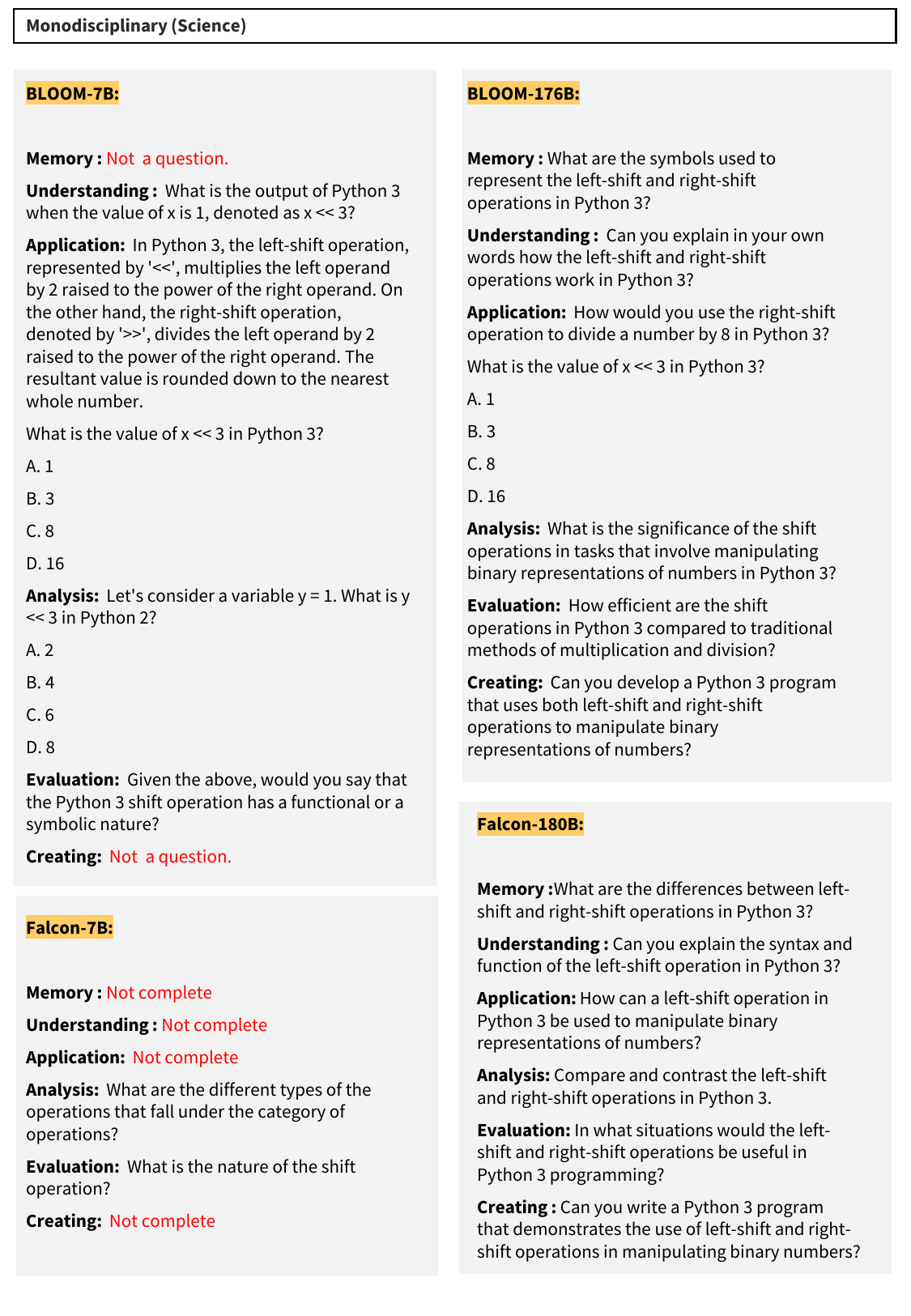}
  \caption{Questions posed by different LLMs in the science-related monodisciplinary domain. }
  \label{fig:3xuexi-case1}
\end{figure*}

\begin{figure*}[!h]
  \centering
  \includegraphics[width=0.8\linewidth]{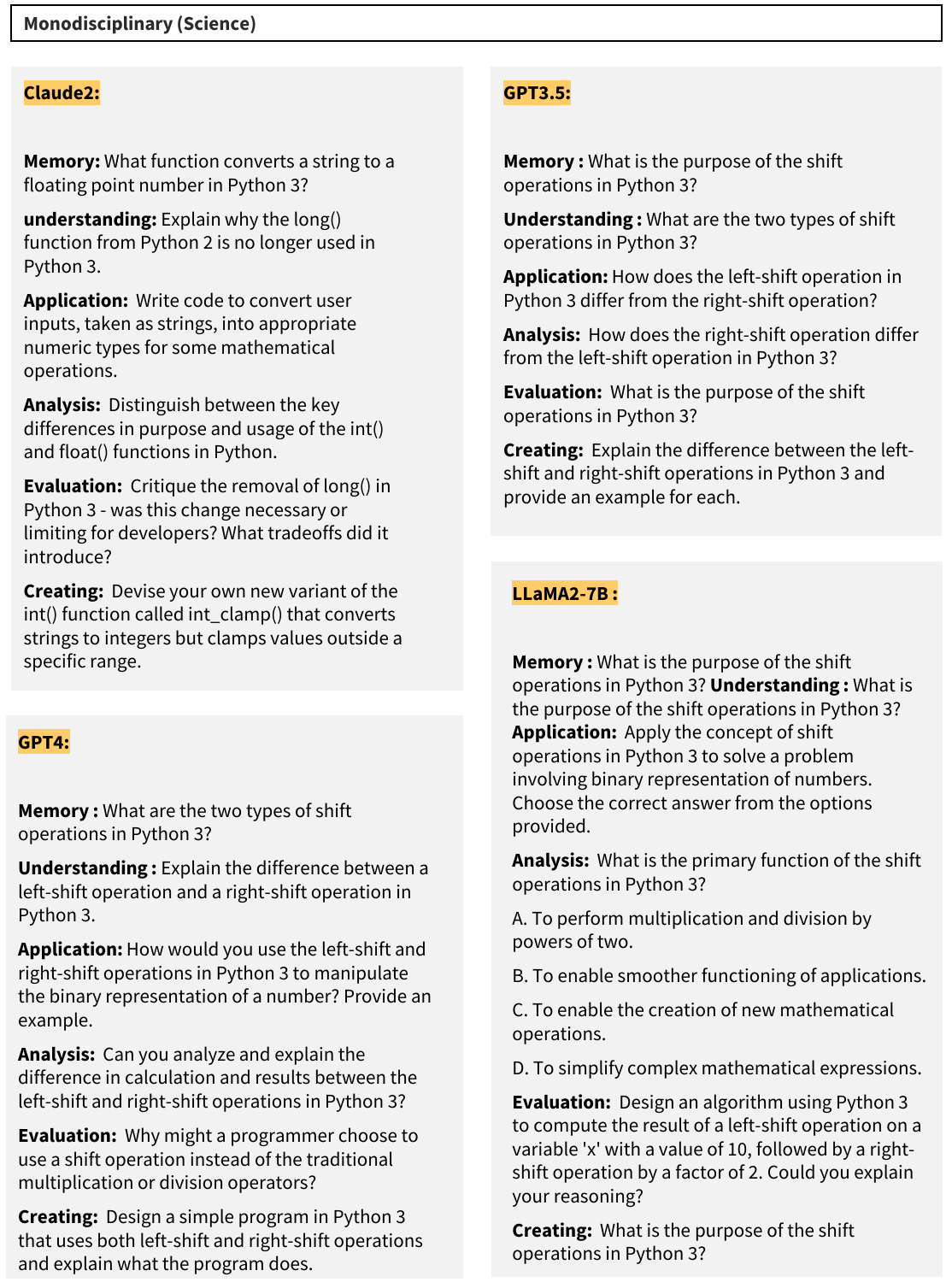}
  \caption{Questions posed by different LLMs in the science-related monodisciplinary domain. }
  \label{fig:3xuexi-case2}
\end{figure*}

\begin{figure*}[!h]
  \centering
  \includegraphics[width=0.8\linewidth]{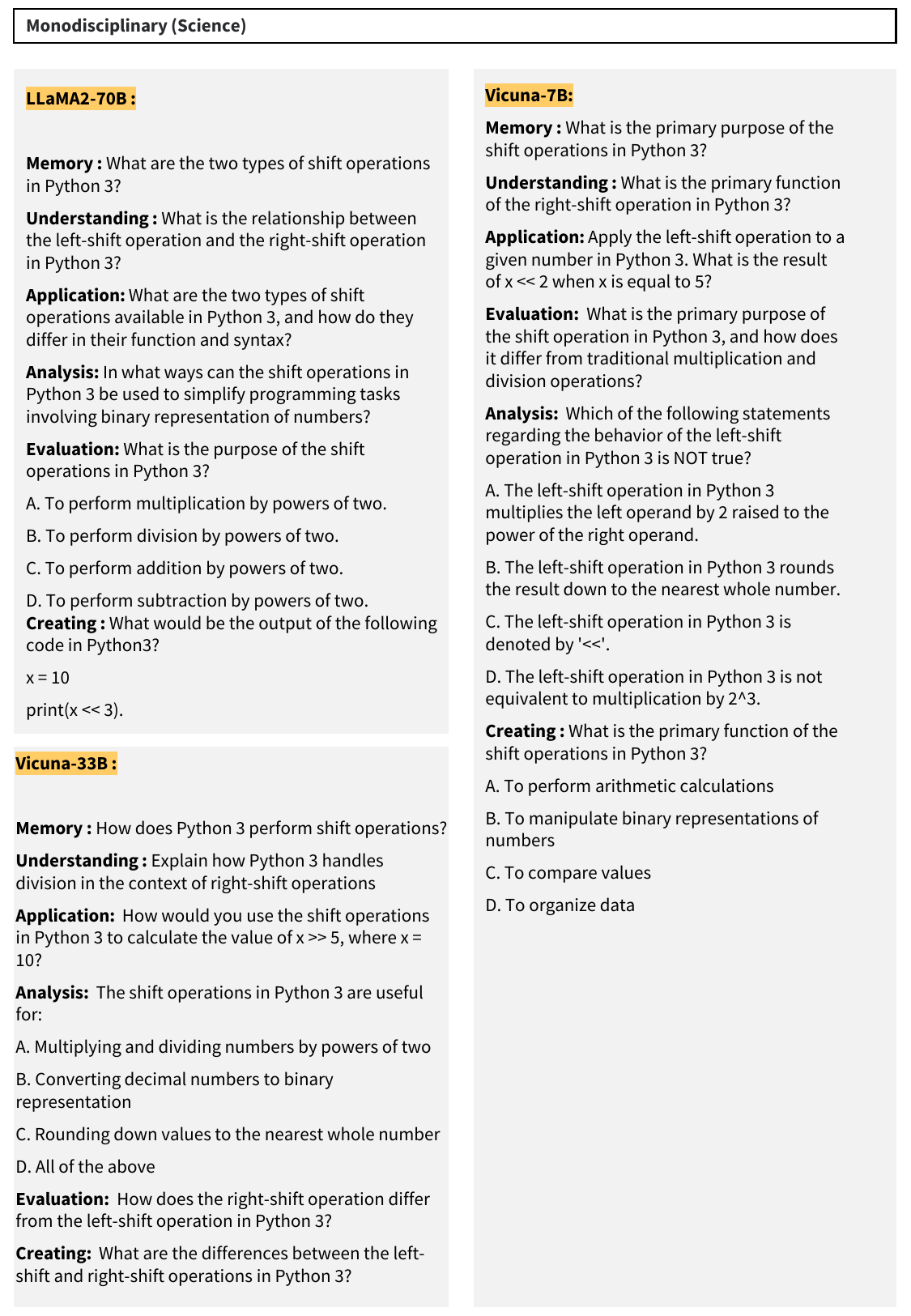}
  \caption{Questions posed by different LLMs in the science-related monodisciplinary domain. }
  \label{fig:3xuexi-case3}
\end{figure*}

\begin{figure*}[!h]
  \centering
  \includegraphics[width=0.8\linewidth]{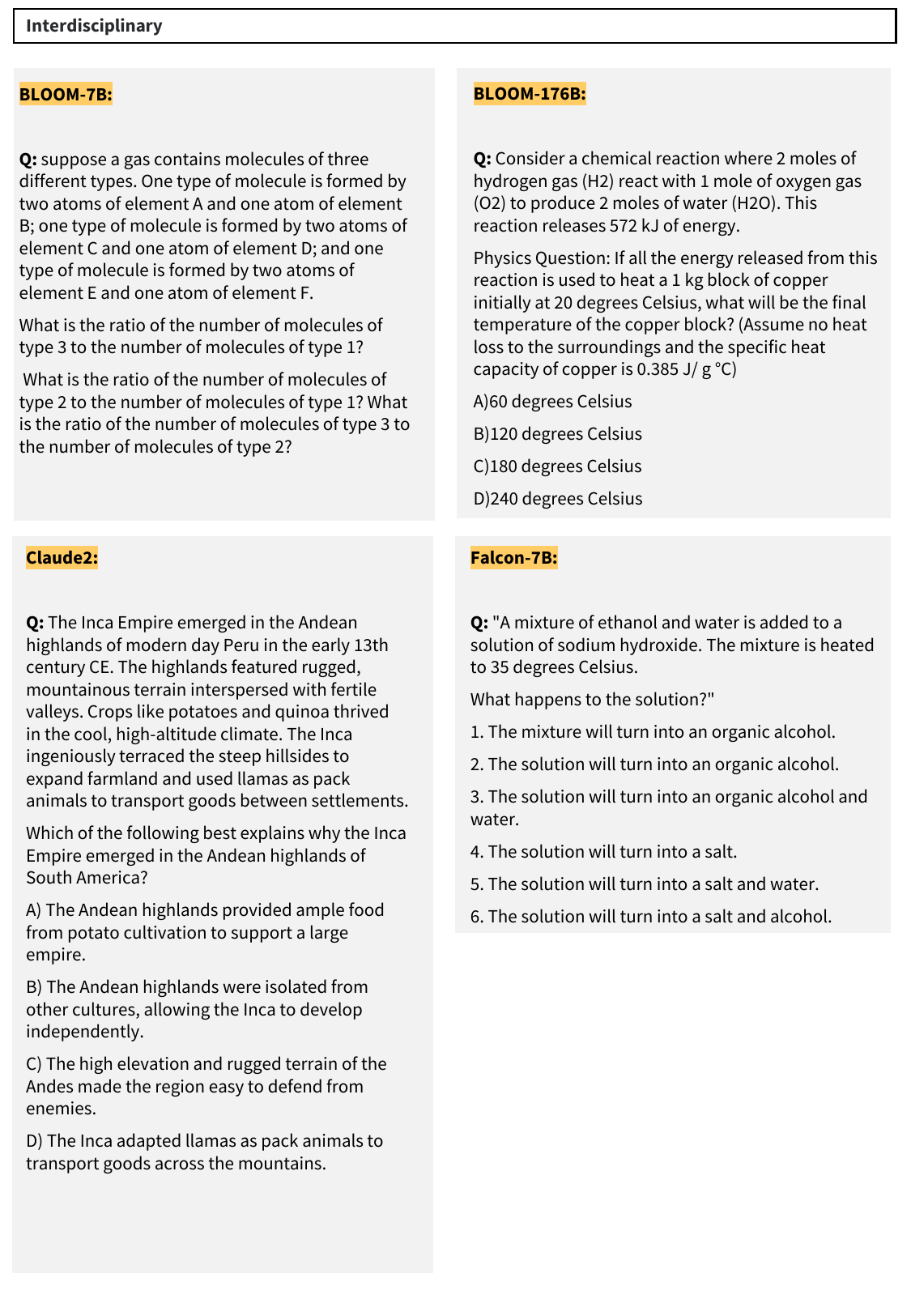}
  \caption{Questions posed by different LLMs in the interdisciplinary domain. }
  \label{fig:4xuexi-case1}
\end{figure*}

\begin{figure*}[!h]
  \centering
  \includegraphics[width=0.8\linewidth]{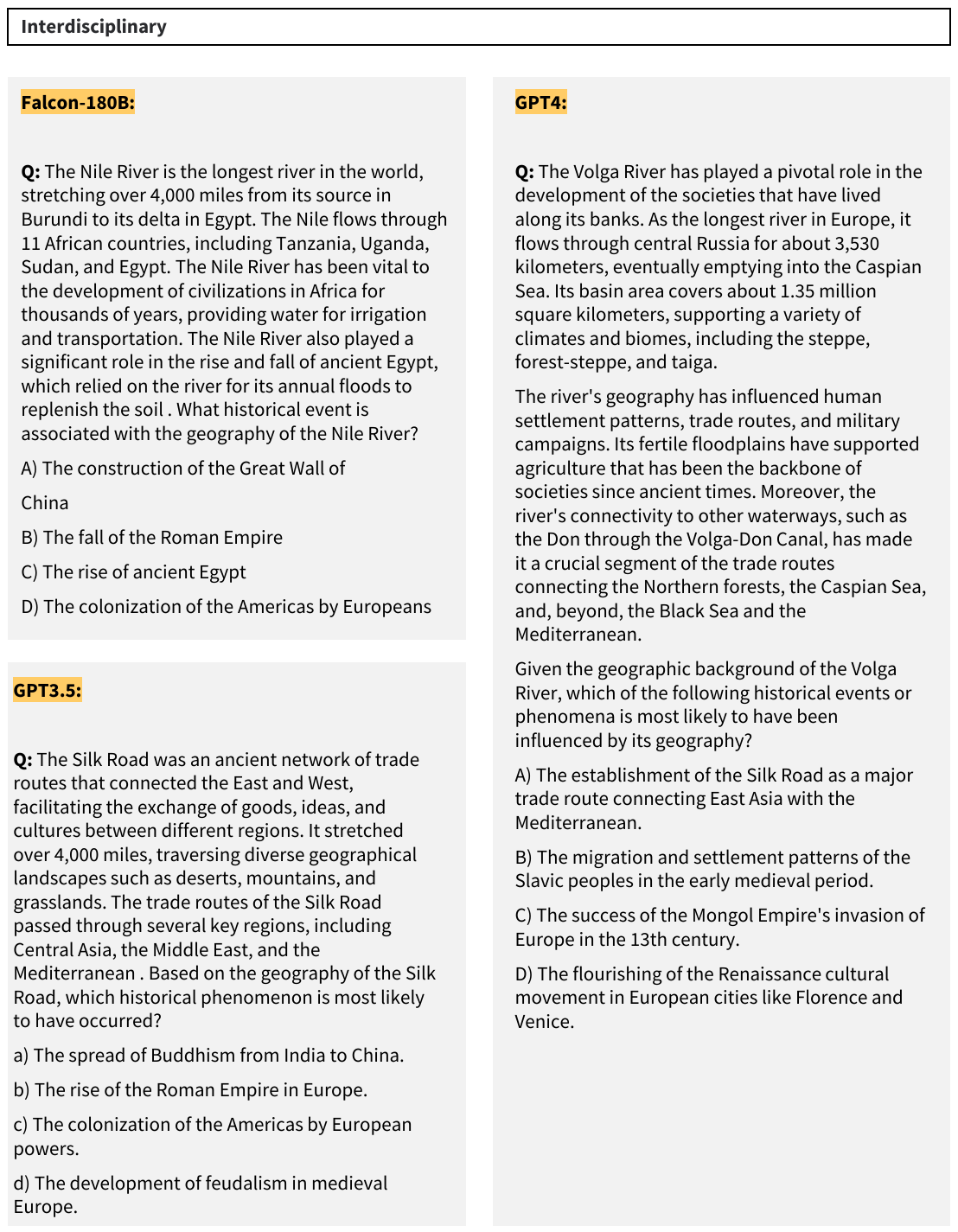}
  \caption{Questions posed by different LLMs in the interdisciplinary domain. }
  \label{fig:4xuexi-case2}
\end{figure*}

\begin{figure*}[!h]
  \centering
  \includegraphics[width=0.8\linewidth]{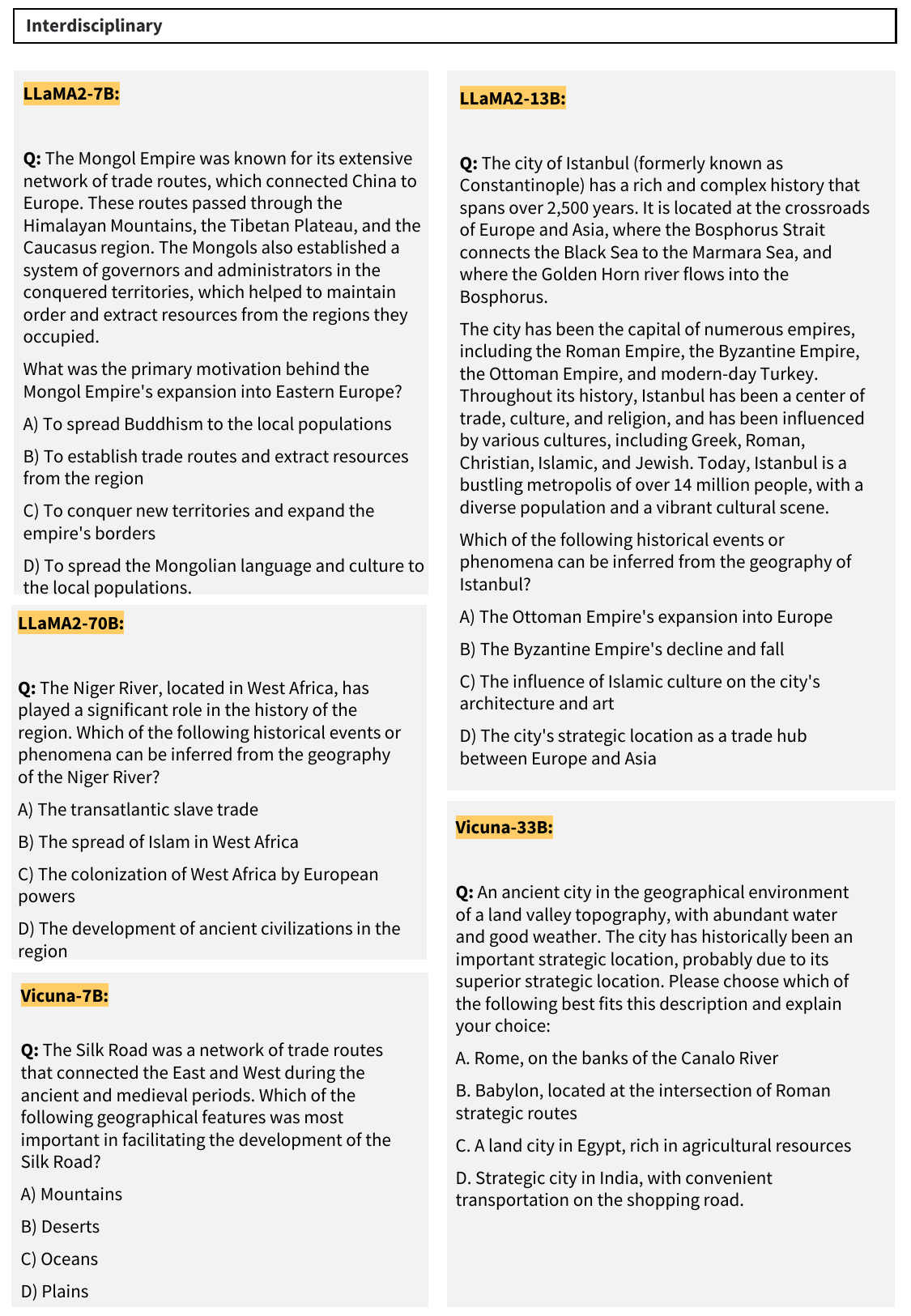}
  \caption{Questions posed by different LLMs in the interdisciplinary domain. }
  \label{fig:4xuexi-case3}
\end{figure*}

\begin{figure*}[!h]
  \centering
  \includegraphics[width=0.8\linewidth]{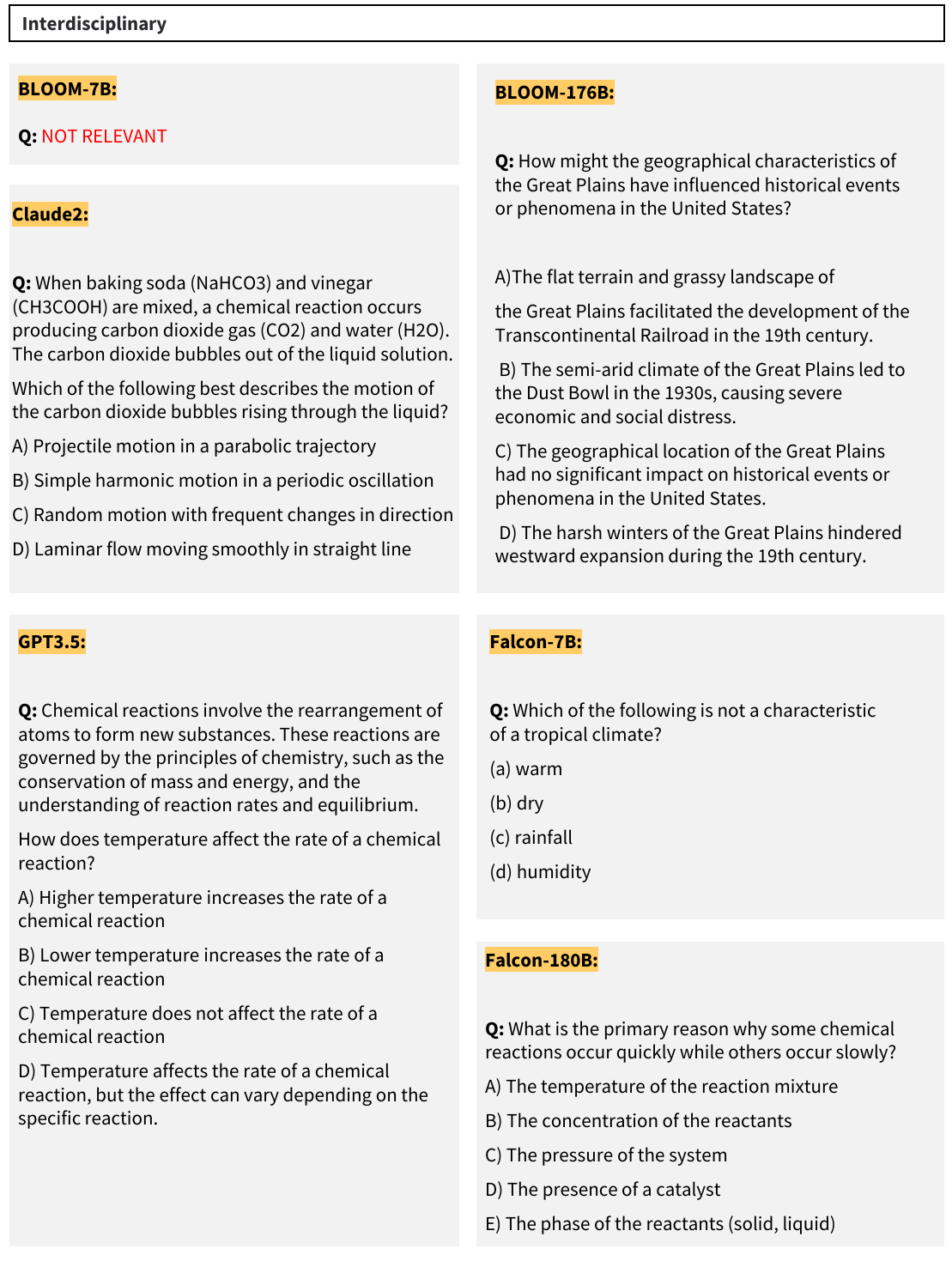}
  \caption{Questions posed by different LLMs in the interdisciplinary domain. }
  \label{fig:4xuexi-case4}
\end{figure*}

\begin{figure*}[!h]
  \centering
  \includegraphics[width=0.8\linewidth]{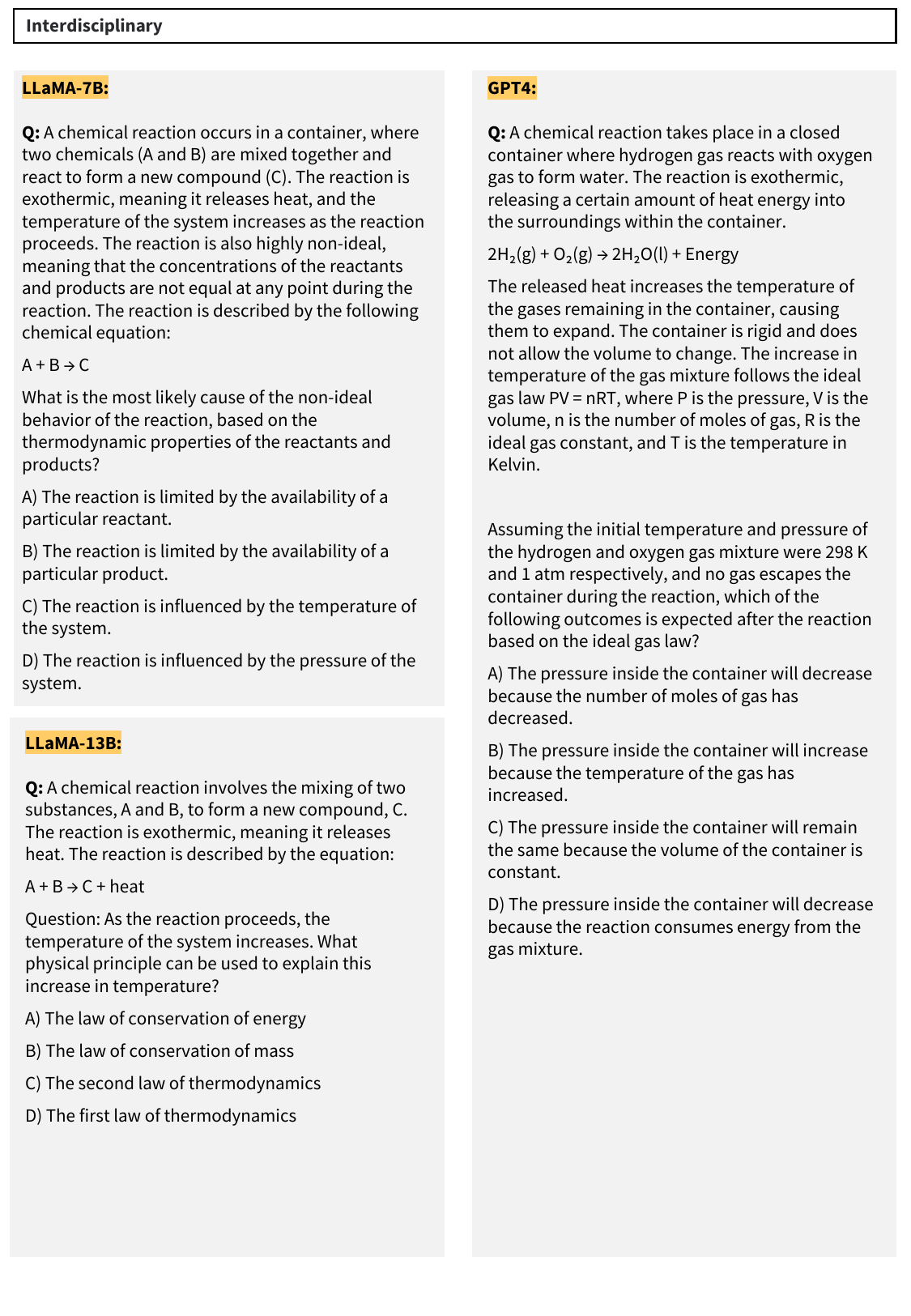}
  \caption{Questions posed by different LLMs in the interdisciplinary domain. }
  \label{fig:4xuexi-case5}
\end{figure*}

\begin{figure*}[!h]
  \centering
  \includegraphics[width=0.8\linewidth]{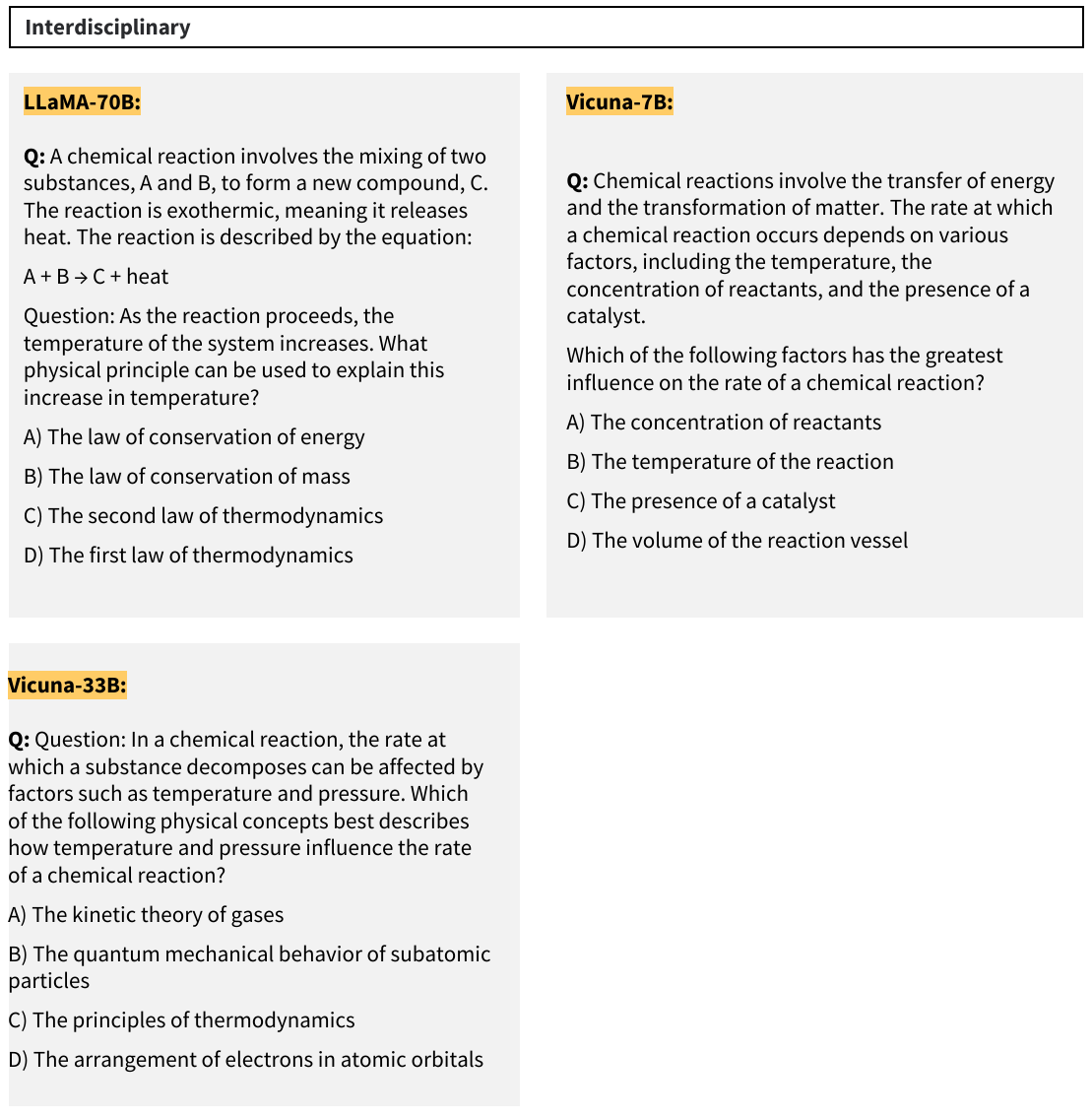}
  \caption{Questions posed by different LLMs in the interdisciplinary domain. }
  \label{fig:4xuexi-case6}
\end{figure*}

\end{document}